\documentclass{article}
\usepackage[preprint]{neurips_2023}
\usepackage[utf8]{inputenc} 
\usepackage[T1]{fontenc}    
\usepackage[dvipsnames]{xcolor}
\usepackage[colorlinks=true, urlcolor=Dandelion ,linkcolor=Dandelion,citecolor=ForestGreen]{hyperref}
\usepackage{url}            
\usepackage{booktabs}       
\usepackage{amsfonts}       
\usepackage{nicefrac}       
\usepackage{svg}            
\usepackage{microtype}      
\usepackage{caption}
\usepackage{ragged2e}
\usepackage[labelfont=bf]{caption}
\usepackage{multirow}
\usepackage{wrapfig}
\usepackage{xspace}
\usepackage{tcolorbox}
\usepackage{graphicx}
\usepackage{subcaption}
\usepackage{fancyhdr,lipsum}
\usepackage[export]{adjustbox}
\usepackage{sectsty}
\usepackage{amssymb}
\usepackage{pifont}
\newcommand{\xmark}{\ding{55}}%
\sectionfont{\color{RoyalBlue}}
\subsectionfont{\color{RoyalBlue}}
\subsubsectionfont{\color{RoyalBlue}}

\tcbuselibrary{most}

\title{\textcolor{ForestGreen}{Tu}\textcolor{Dandelion}{ca}\textcolor{RoyalBlue}{no}: Advancing Neural Text Generation for Portuguese}

\author{                    
    Nicholas Kluge Corrêa
    \And
    Aniket Sen
    \And
    Sophia Falk
    \And
    Shiza Fatimah
    \vspace{5mm}\\
    \textbf{Rhenish Friedrich Wilhelm University of Bonn}\\ 
}

\begin{document}
\maketitle

\begin{figure}[h]
\centering
\includegraphics[width=0.40\linewidth]{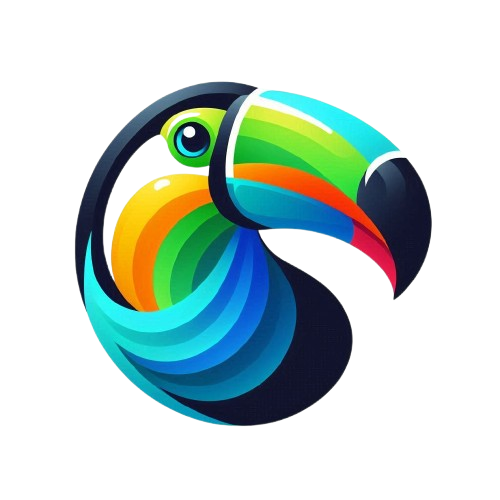}
\end{figure}

\begin{abstract}

Significant advances have been made in natural language processing in recent years. However, our current deep learning approach to language modeling requires substantial resources in terms of data and computation. One of the side effects of this data-hungry paradigm is the current schism between languages, separating those considered high-resource, where most of the development happens and resources are available, and the low-resource ones, which struggle to attain the same level of performance and autonomy. This study aims to introduce a new set of resources to stimulate the future development of neural text generation in Portuguese. In this work, we document the development of \textbf{\textcolor{ForestGreen}{Gig}\textcolor{Dandelion}{aVe}\textcolor{RoyalBlue}{rbo}}, a concatenation of deduplicated Portuguese text corpora amounting to 200 billion tokens. Via this corpus, we trained a series of decoder-transformers named \textbf{\textcolor{ForestGreen}{Tu}\textcolor{Dandelion}{ca}\textcolor{RoyalBlue}{no}}. Our models perform equal or superior to other Portuguese and multilingual language models of similar size in several Portuguese benchmarks. The evaluation of our models also reveals that model performance on many currently available benchmarks used by the Portuguese NLP community has \textit{little to no correlation} with the scaling of token ingestion during training, highlighting the limitations of such evaluations when it comes to the assessment of Portuguese generative language models. All derivatives of our study are openly released on \href{https://github.com/Nkluge-correa/Tucano}{GitHub}\footnote{\hspace{1mm}\includegraphics[scale=0.025]{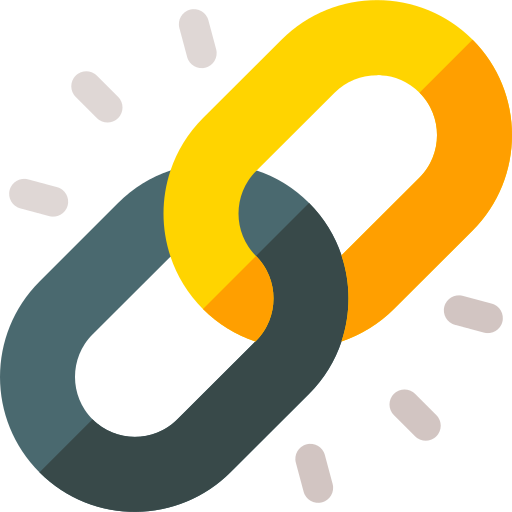}\hspace{1mm} \href{https://github.com/Nkluge-correa/Tucano}{github.com/Nkluge-correa/Tucano}} and \href{https://huggingface.co/TucanoBR}{Hugging Face}\footnote{\hspace{1mm}\includegraphics[scale=0.025]{img/link.png}\hspace{1mm} \href{https://huggingface.co/TucanoBR}{huggingface.co/TucanoBR}}.

\end{abstract}

\section{Introduction}

Over almost a decade, the deep learning paradigm has been the \textit{de facto} mode of operation for many of the sub-fields involved in artificial intelligence research \cite{lecun2015deep, goodfellow2016deep}. Natural Language Processing (NLP) is a canonical depiction of the success story of deep learning \cite{nadkarni2011natural, deng2018deep, otter2020survey}, where neural network approaches to machine learning have become the engine that powers many aspects of our current age of intelligent automation, with breakthroughs like word embeddings \cite{mikolov2013linguistic, mikolov2013efficient} and the transformer neural architecture \cite{vaswani2017attention} being at the heart of this revolution.

Another aspect of this developmental movement is using the self-supervised learning approach as an intermediate step to many language modeling tasks \cite{geiping2023cookbook}. In essence, self-supervised learning is a training methodology for machine learning systems, where we leverage the vastness of available unlabeled data at our disposition to create pretraining tasks where labeling can happen on the fly. This results in systems with useful and downstream-applicable representations tied to the domain they were trained on \cite{hastie2009overview, misra2020self, geiping2023cookbook}. This training approach has been responsible for some of the early breakthroughs of the field \cite{bengio2000neural, mikolov2013efficient, sutskever2014sequence, bahdanau2014neural}, which have now morphed into our standard training recipe for foundation models \cite{bommasani2021opportunities}.

Nonetheless, while the statement "\textit{leverage the vastness of available unlabeled data at our disposition to create pretraining tasks}" can be true for languages like English or Chinese, where datasets can reach the $10^{13}$ tokens mark \cite{dolma, young2024yi, chen2023chinesewebtext, penedo2024finewebdatasetsdecantingweb}, and models are trained way past what scaling laws prescribe as compute-optimal \cite{zhang2024tinyllama, parmar2024nemotron, xue2024openmoe}, the same cannot be said about the crushing majority of more than 7000 languages spoken around the world today \cite{rodrigues2023advancing, lopes2024gl, cohere2024gap}. Hence, the prospect of training language models at the scale required to match what is done in such high-resource languages (even when compared to the state-of-the-art from 5 years ago)\footnote{Models like GPT-3 (2020) \cite{brown2020language} were trained only on 300B tokens, which is still something not reproduced in a monolingual setting for most low-resource languages.} is a far-fetched goal for most low-resource languages \cite{gandhe2014neural, adams2017cross, cruz2019evaluating, muennighoff2023scaling, correa2024teenytinyllama}.

To overcome this linguistic shortcoming, one of the approaches found in the literature is the development of multilingual models and datasets \cite{singh2024ayadatasetopenaccesscollection, ustun2024ayamodelinstructionfinetuned, aryabumi2024aya, srivastava2024lolaopensourcemassively}. In these models, the self-supervised pretraining stage is conducted with various languages. Models like  mBERT \cite{devlin2018bert}, mT5 \cite{xue2020mt5}, XLM-RoBERTa \cite{conneau2020unsupervised}, mGPT \cite{shliazhko2022mgpt},  XGLM \cite{lin2021few}, BLOOM \cite{workshop2022bloom}, PolyLM \cite{wei2023polylm}, Aya \cite{ustun2024ayamodelinstructionfinetuned}, and Llama 3 \cite{dubey2024llama} are examples of this approach. On the other hand, the development of monolingual language models has also been explored and, at many times, shown to be a more successful approach to the multilingual one, like in the case of Finish \cite{virtanen2019multilingual}, French \cite{martin-etal-2020-camembert}, Catalan \cite{armengol2021multilingual}, Chinese \cite{sun2021ernie}, and Portuguese \cite{souza2020bertimbau, rodrigues2023advancing, correa2024teenytinyllama}. Besides, as already pointed out by other works \cite{correa2024teenytinyllama}, if based on raw pretraining instead of a fine-tuning approach, the monolingual approach can help developers escape the computational (i.e., models that are too expensive to run) and legal constraints (i.e., models that are restricted in terms of their licensing) of working with an already established foundation.

However, advances in developing low-resource monolingual language models, such as those for Portuguese, remain limited, small in scale, undocumented, lacking standardization, and often reliant on repurposing models trained behind closed doors,\footnote{This is particularly true for the European and Brazilian variants, with other variants (e.g., Angolan Portuguese) even less represented or entirely absent.} as will be discussed in the next section. These deficits also make it challenging to compare language models and evaluation benchmarks. At the same time, the effectiveness of the currently available benchmarks for Portuguese is also untested. In this work, we aim to address these challenges and build on existing studies to improve the status of generative language modeling research and development for Portuguese. In summary, our study offers the following advancements to the Portuguese NLP community:

\begin{enumerate}
    \item The concatenation of a larger and more high-quality dataset for Portuguese language modeling (\textbf{\textcolor{ForestGreen}{Gig}\textcolor{Dandelion}{aVe}\textcolor{RoyalBlue}{rbo}}).
    \item The development of learned filters and datasets to improve text pre-processing for Portuguese.
    \item Pushing self-supervised pretraining beyond the 500B tokens mark for Portuguese monolingual models.
    \item The development of new, low-resource, efficient, and effective open-source foundation models for Portuguese (\textbf{\textcolor{ForestGreen}{Tu}\textcolor{Dandelion}{ca}\textcolor{RoyalBlue}{no}}).
    \item A critical assessment and comparison of currently available benchmarks for Portuguese language models.
\end{enumerate}

In Section \ref{section2}, we review the current status of Portuguese Large Language Model (LLM) research and development, documenting the trends and deficits in the field. Section \ref{section3} describes the pretraining corpus used in this work. Section \ref{section4} and \ref{section5} contain the definition of our chosen tokenizer and the parameter space of different models we trained. In Section \ref{section6}, we discuss the training and evaluation of our models. We also employ a simple alignment strategy to our more capable models, as will be discussed in Section \ref{section7}. In Section \ref{section8}, we present the results of our evaluation harness. Finally, Sections \ref{section9} and \ref{section10} provide an outlook for future studies and conclusion of our work.

\section{An Anthology of Portuguese LLM Development}
\label{section2}

A historical timeline of Portuguese LLM research and development can help to understand how our work should be contextualized. This landscape consists of many pre-trained and fine-tuned transformer networks. However, before doing so, we would like to differentiate between two terms (\textit{fine-tuning} and \textit{pretraining}) that are used loosely and interchangeably in the literature, making it sometimes difficult to distinguish between them.

First, we will use the definition of fine-tuning as \textit{"the process of updating the weights of a pre-trained model on new data"} \cite{goodfellow2016deep, prince2023understanding}. Hence, all models that adopt a training methodology in which already trained weights are repurposed and updated are byproducts of a fine-tuning approach, done at full or low ranks \cite{hu2021lora}, with or without adaptations (e.g., changing the tokenizer vocabulary and re-initializing the embedding matrix and language modeling head). Secondly, pretraining can be defined as \textit{"the act of training a neural network initialized with random weights"}. The distinction between pretraining and training is merely contextual or terminological, given that a foundation model is usually trained to be later "trained again" (i.e., fine-tuned) for a more specific task, hence the "pre", as "before we train on the tasks and applications we care about". Although intuitive, this distinction is sometimes presented in an unclear fashion, even though the difference between both approaches is evident and can severely affect the performance models can achieve.\footnote{For example, the fine-tuning of a foundation like Llama 2 \cite{touvron2023llama2}, even on a small dataset of monolingual text, usually results in a model that, besides presenting performant language modeling skills in the language chosen, inherits from the pervasive training the original foundation was put through. However, this approach can make results hard to trace depending on the context in which we find ourselves. For example, how much of performance $x$ can be attributed to $10^{13}$ tokens of pretraining compared to the $10^{9}$ tokens of fine-tuning in benchmark $y$? Such a question becomes even harder to answer when models are based on foundations developed in a private setting.}

In summary, distinguishing between these two developmental approaches is essential to interpret evaluation results and model capabilities, as we will explore in upcoming sections. Now, with these definitions in mind, let us review some of the developments achieved in recent years (Fig. \ref{fig:timeline}):

\begin{figure}
    \centering
    \includegraphics[width=400pt]{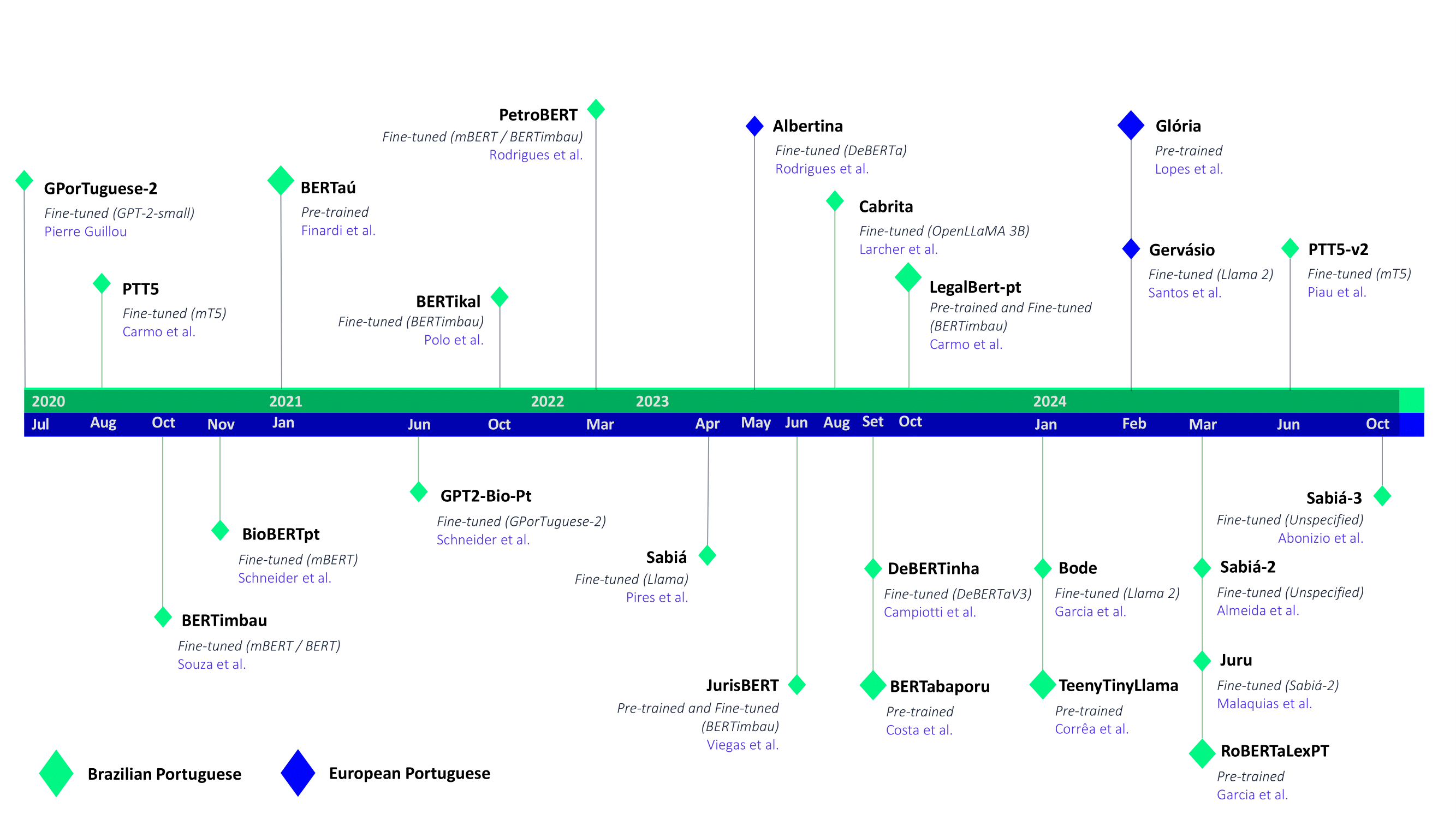}
    \caption{This timeline illustrates several Portuguese language model releases from 2020 to October 2024. The models are color-coded to indicate their respective Portuguese language variants, e.g., green for South America and blue for Europe. The timeline also distinguishes pre-trained models from fine-tuned derivatives of other foundations. We limited the models displayed in this timeline to those we could find tied to publication reports, unpublished manuscripts, peer-reviewed papers, and popular repositories.}
    \label{fig:timeline}
\end{figure}

\begin{itemize}

    \item \textbf{GPorTuguese-2 (July 18, 2020) \cite{pierre2020gpt2smallportuguese}}: The first publicly available large language model tailored for Brazilian Portuguese. GPorTuguese-2 is a byproduct of fine-tuning OpenAI's smallest version of GPT-2 \cite{radford2019language} on the Portuguese portion of Wikipedia \cite{wikidump}. This model also has adaptations, like its own byte-pair encoding (BPE) tokenizer with a custom vocabulary that repurposes the joint embeddings from the original English vocabulary. GPorTuguese-2 was fine-tuned on $\approx$ 1.2 GB of text, and it is available under an MIT License.\\

    \item \textbf{PTT5 (August 20, 2020) \cite{carmo2020ptt5}}: An encoder-decoder model developed as a foundation for Text-to-Text tasks in Brazilian Portuguese. PTT5 is an adapted version of another foundation model (Google's multilingual T5 \cite{raffel2020exploring}), having a custom vocabulary and embeddings that were reinitialized and trained from scratch. PTT5 model was trained on the BrWaC corpus \cite{wagner2018brwac} ($\approx$ 2.68 billion tokens) and is available under an MIT License.\\

    \item \textbf{BERTimbau (October 20, 2020) \cite{souza2020bertimbau}}: A fine-tuning version of the base and large versions of mBERT and English BERT \cite{devlin2018bert}, respectively. BERTimbau has a custom vocabulary, embeddings, and attention heads that were reinitialized and trained from scratch. Both versions of BERTimbau were trained on BrWaC \cite{wagner2018brwac}, and are available under an MIT License.\\

    \item \textbf{BioBERTpt (November 19, 2020) \cite{rubel2020biobertpt}}: A fine-tuned version of mBERT \cite{devlin2018bert}. BioBERTpt was created to support named-entity recognition (NER) in clinical and biomedical applications, being trained on a corpus of 44.1 M tokens of clinical narratives and biomedical-scientific papers in Brazilian Portuguese. While mBERT is licensed under an MIT License, BioBERTpt does not specify any licensing regime. However, the model is openly accessible via the Hugging Face platform.\\

    \item \textbf{BERTaú (January 28, 2021) \cite{finardi2021berta}}: A pre-trained BERT-based LLM for Brazilian Portuguese. BERTaú was pre-trained using customer-service conversations from a Brazilian financial services company (Itaú), with 5GB of text, following a similar training protocol to the one described in the original BERT paper \cite{devlin2018bert}. As far as we could investigate, BERTaú is not open to the public, being proprietary software from Itaú.\\

    \item \textbf{GPT2-Bio-Pt (June 1, 2021) \cite{schneider2021gpt}}: A fine-tuned version of the GPorTuguese-2 \cite{pierre2020gpt2smallportuguese}, trained on 48M tokens of clinical and biomedical literature. While GPorTuguese-2 is licensed under an MIT License, GPT2-Bio-Pt does not specify any licensing regime. However, the model is accessible via the Hugging Face platform.\\

    \item \textbf{BERTikal (October 5, 2021) \cite{polo2021legalnlp}}: A BERT model tailored for the Brazilian Portuguese legal domain. BERTikal is a fine-tuned version of BERTimbau-base \cite{souza2020bertimbau}. For training, the authors used 2.6 GB of text composed of legal documents from several Brazilian courts dated from 2019 to 2020. BERTikal is currently available under an MIT License.\\

    \item \textbf{PetroBERT (March 16, 2022) \cite{rodrigues2022petrobert}}: A BERT-based model adapted to the oil and gas exploration domain in Portuguese. PetroBERT has two versions, each fine-tuned over a different foundation: mBERT \cite{devlin2018bert} and BERTimbau \cite{souza2020bertimbau}. No model is currently available for public use.\\

    \item \textbf{Sabiá (April 16, 2023) \cite{pires2023sabi}}: A series of fine-tuned models that used GPT-J \cite{gptj} and Llama \cite{touvron2023llama1} as a foundation. The outcomes of this fine-tuning process are Sabiá-7b, 65B (both derivatives of Llama), and Sabiá-J (using GPT-J as a base). The Sabiá series was trained on $\approx$ 7.8 billion tokens from a filtered portion of the ClueWeb 2022 dataset \cite{overwijk2022clueweb22}. Sabiá-65B and Sabiá-J are unavailable to the public, while Sabiá-7B is available under the Llama 2 license.\\

    \item \textbf{Albertina (May 11, 2023) \cite{rodrigues2023advancing}}: A family of encoder-only transformers that use DeBERTa \cite{he2020deberta} as a foundation. Albertina models come for Brazilian and European Portuguese, having been trained on over 2.2B tokens of text. Currently, the Albertina series scales from 100 million to 1.5 billion parameters, and all models are available under an MIT license.\\

    \item \textbf{JurisBERT (July 30, 2023) \cite{viegas2023jurisbert}}: A series of BERT-based models developed for the Brazilian legal domain. In this series, we find models either pre-trained from scratch or adapted from BERTimbau-base \cite{souza2020bertimbau}. We also find adapted versions from these models that were later fine-tuned to work as Sentence Transformers \cite{reimers2019sentence}. Even though no license is tied to these models, all are available for use in the Hugging Face platform.\\

    \item \textbf{Cabrita (August 23, 2023) \cite{larcher2023cabrita}}: A fine-tuned version of OpenLLaMA 3B \cite{openlm2023openllama}, with an adapted tokenizer and extended embeddings. Cabrita was trained on 7 billion tokens extracted from the Portuguese subset of the mC4 dataset \cite{xue2020mt5}. Cabrita is available under an Apache 2.0 license.\\

    \item \textbf{BERTabaporu (September 4, 2023) \cite{costa2023bertabaporu}}: Two BERT models, base and large, pre-trained on Brazilian Portuguese Twitter data. These models were trained on 2.9B tokens, following a similar training recipe as the original BERT paper \cite{devlin2018bert}. BERTabaporu is available under an MIT license.\\

    \item \textbf{DeBERTinha (September 28, 2023) \cite{campiotti2023debertinha}}: An adapted version of DeBERTaV3 \cite{he2021debertav3}, fine-tuned to be performant in Brazilian Portuguese. DeBERTinha has a custom vocabulary and embeddings trained from scratch while repurposing the other weights from the original DeBERTaV3. For training, the authors used a combination of the BrWaC \cite{wagner2018brwac} and Carolina \cite{corpusCarolinaV1} datasets, which amounted to 33GB of text ($\approx$ 3.4 billion tokens). DeBERTinha is available under an MIT license.\\

    \item \textbf{LegalBert-pt (October 12, 2023) \cite{silveira2023legalbert}}: Both a pre-trained BERT and a fine-tuned BERTimbau \cite{souza2020bertimbau}. The training dataset contained 1.5 million samples of legal texts (12 million sentences) and was used to pre-train/fine-tune both versions of LegalBert-pt. Both versions of LegalBert-pt are available under the OpenRAIL license.\\

    \item \textbf{Bode (January 5, 2024) \cite{bode2024}}: Both a low-rank adaptation and full fine-tuned version of Llama 2 \cite{touvron2023llama2}. These models were trained on a translated version of the Alpaca dataset \cite{taori2023alpaca} (i.e., 52,000 instruction-following demonstrations generated by text-davinci-003). Bode is available in two sizes, 7B and 13B, under the Llama 2 license.\footnote{Similar models (e.g., Caramelo \cite{henrique2023caramelo} and Harpia \cite{henrique2023harpia}) can also be found using Falcon 7B as a foundation \cite{almazrouei2023falcon}.}\\

    \item \textbf{TeenyTinyLlama (January 30, 2024) \cite{correa2024teenytinyllama}}: A pair of language models pre-trained in Brazilian Portuguese. TeenyTinyLlama (TTL) models are based on the Llama architecture \cite{touvron2023llama2}, downsized to a 160 and 460 million parameter version. These were trained on a concatenation of publicly available Portuguese datasets called Portuguese-Corpus Instruct (6.2B tokens). Models, datasets, and source code for training/evaluation are available under an Apache 2.0 license.\\

    \item \textbf{Glória (February 20, 2024) \cite{lopes2024gl}}: A pair of language models pre-trained in European Portuguese. Glória models are based on the GPTNeo architecture \cite{black2022gpt}, scaled to 1.3B and 2.7B parameters. Its training dataset comprises a concatenation of European Portuguese datasets amounting to 35.5 billion tokens. Glória's usage is restricted to research-only purposes, subject to the ClueWeb22 Dataset license.\\

    \item \textbf{Gervásio (February 29, 2024) \cite{santos2024advancing}}: A fined-tuned version of Llama 2 7B \cite{touvron2023llama2}. It comes in a European and Brazilian variant, each trained on distinct datasets designed to induce instruction-following behavior. Even though Gervásio is a derivative of Llama 2,\footnote{Hence, supposedly should be restricted by the \href{https://ai.meta.com/llama/license/}{Llama 2 Community License Agreement}.} Gervásio is currently available under an MIT license.\\

    \item \textbf{RoBERTaLexPT (March 12, 2024) \cite{garcia2024robertalexpt}}: A pair of encoder-only LLM based on the RoBERTa-base implementation \cite{liu2019roberta}, tailored for general Brazilian Portuguese language modeling and applications in the legal domain. While RoBERTaCrawlPT was pre-trained from scratch on the CrawlPT corpora (i.e., a deduplicated concatenation of BrWaC \cite{wagner2018brwac}, Common Crawl \cite{wenzek-etal-2020-ccnet, conneau-etal-2020-unsupervised}, and Oscar \cite{OrtizSuarezSagotRomary2019, ortiz-suarez-etal-2020-monolingual, abadji2022towards}), RoBERTaLexPT was pre-trained from scratch from a combination of CrawlPT and LegalPT (i.e., a concatenation of six different Brazilian Portuguese legal text corpora \cite{niklaus2023multilegalpile, rodrigues2023advancing, sousa2019iudicium, bonifacio2020study}). RoBERTaCrawlPT and RoBERTaLexPT are available under a Creative Commons license (CC BY 4.0).\\

    \item \textbf{Sabiá-2 (March 14, 2024) \cite{almeida2024sabi}}: Not much information is known about Sabiá-2, and its report only brings evaluation scores of internally held benchmarking on two models of unknown sizes, referred to by the authors as "small" and "medium". Sabiá-2 is only available to the public via a commercial API.\\

    \item \textbf{Juru (March 26, 2024) \cite{junior2024juru}}: A fine-tuned version of Sabiá-2 small. Juru was trained on 5.88 billion tokens from academic studies and other high-quality sources tied to the Brazilian legal domain. Juru and the dataset used to train it are not available to the public. \\

    \item \textbf{PTT5-v2 (June 16, 2024) \cite{piau2024ptt5}}: Similar to the first iteration of PTT5, PTT5-v2 is a series of fine-tuned models, up to 3 billion parameters, based on Google's multilingual T5 \cite{raffel2020exploring}. PTT5-v2 was trained on approximately 524 GB of uncompressed text for 1.7 million optimization steps (115 billion tokens), following a training regime similar to the original T5 paper. Even though no license is tied to these models, all are available for use in the Hugging Face platform. \\

    \item \textbf{Sabiá-3 (October 15, 2024)}: Not much information is known about Sabiá-3, and its report only brings evaluation scores of internally held benchmarking on one model of unknown size. Sabiá-3 is only available to the public via a commercial API.\footnote{The "\href{https://www.maritaca.ai/_files/ugd/6cb9d6_73c5960d94c44b09ba4daf8037f7003a.pdf}{Sabiá-3 Technical Report"} is available via the maritaca.ai website.}\\

\end{itemize}

Reviewing these past works reveals a few crucial insights about Portuguese NLP research's current state and direction. Firstly, language adaptation, i.e., repurposing the language modeling capabilities of a model for another language, is a popular approach, especially when the foundation used is already performant and multilingual. The great majority of the work mentioned in the above list presents research revolving around fine-tuning and adaptation of already pre-trained models rather than developing native Portuguese foundations \cite{souza2020bertimbau, rodrigues2023advancing, larcher2023cabrita, almeida2024sabi}.

Moreover, the fine-tuning over pretraining choice can be attributed to factors characteristic of low-resource languages (e.g., not enough tokens) and conditions of low-resource development (e.g., not enough computing). For example, until 2024, almost all studies were limited to datasets with less than 10 billion tokens, with most fine-tuning models using much less than this. Although justifiable in terms of model size and scaling laws \cite{hoffmann2022training}, this makes the training of larger models infeasible unless we promote a severe repetition of our dataset \cite{larcher2023cabrita}, which, however, does not contribute to improved model performance \cite{muennighoff2023scaling}. Hence, most of the community relies heavily on leveraging the capabilities of established models, which have been extensively pre-trained on large and diverse datasets in other languages.

Another interesting point is that while encoder-only models like BERT have dominated the landscape for some time \cite{souza2020bertimbau, polo2021legalnlp, rodrigues2022petrobert, rodrigues2023advancing, costa2023bertabaporu, silveira2023legalbert, campiotti2023debertinha, garcia2024robertalexpt}, there has been a recent shift towards training decoder-only models \cite{pires2023sabi, correa2024teenytinyllama, lopes2024gl}. However, a significant challenge with these models is the need for more standardization in evaluation protocols. Each study tends to develop its own benchmarks and evaluation metrics, which complicates direct comparisons and makes it difficult to ascertain the actual performance of these models. Furthermore, many of the current Portuguese benchmarks available for the evaluations of few-shot capabilities of generative models are either repurposed datasets initially created for downstream development or assessment of BERT-style models \cite{real2020assin, vargas2022hatebr, brum2017building} or translated versions of English benchmarks \cite{lai2023okapi}, which raises questions regarding their effectiveness in evaluating the capabilities of generative language models. Meanwhile, model comparisons are still very limited among Portuguese language models, given that only a few available models allow cheap and accessible benchmarking for cross-study comparisons.

Regarding dataset creation, there is a notable trend towards concatenating and deduplicating various text corpora to form more extensive and scalable datasets. In 2024, we see several studies implementing this approach, giving birth to some of the first large datasets (> 10B tokens) for Portuguese language modeling \cite{lopes2024gl, garcia2024robertalexpt}. However, data filtering and preprocessing methods remain primarily heuristic (e.g., hash-similarity-based deduplication, HTML removal, and mojibake correction) in most studies \cite{lopes2024gl, garcia2024robertalexpt}. At the same time, works that pioneer the filtering and creation of high-quality text datasets do not make these available for the community \cite{junior2024juru, piau2024ptt5}. Meanwhile, we currently do not see the documented use of more sophisticated approaches (e.g., "LLM-as-a-Judge" or learned filters \cite{gunasekar2023textbooks}) to ensure data quality in the creation of these corpora.

It is also worth noting that several recent works have demonstrated the advantages of pretraining models from scratch over fine-tuning/adapting existing ones \cite{costa2023bertabaporu, correa2024teenytinyllama, lopes2024gl, garcia2024robertalexpt}, especially in circumstances where training data is sufficient. Nonetheless, the top-performing models in the literature often rely on fine-tuning foundations whose pretraining data is not disclosed \cite{pires2023sabi, almeida2024sabi}. This opacity raises questions about the factors driving their performance and the limits of how far we can push Portuguese pretraining natively. This brings us to another crucial insight: a significant need for more openness regarding datasets and code implementations across many works. Without it, machine learning research becomes vulnerable to several criticisms, aggravating its current reproducibility crisis \cite{kapoor2022leakage}, while also fueling the "deep learning is alchemy" critique \cite{hutson2018has}.

Finally, another idea worth expressing regarding the pretraining versus fine-tuning choice is that, while building on top of ready-made foundations has its merits (e.g., simplifying the LLM development process to a transfer learning/fine-tuning problem), it is also responsible for masking or diverging attention from severe issues many NLP researchers face. For example, if we agree that LLMs are valuable tools, should communities of low-resource languages be forever bound to \textit{"wait and recycle"} the outputs of research often done behind closed doors and with no prospect of accurate reproducibility? Suppose \textit{"yes"} is the answer. In that case, there is an argument to be made that many communities involved in NLP research find themselves bound in a form of technological colonialism.\footnote{Technological colonialism refers to the dominance of a small number of entities, typically large corporations or specific geographic regions, in controlling and shaping the development, deployment, and norms of advanced technological systems \cite{arnold2005europe}.} On the contrary, if technological sovereignty should be sought as something that "ought", research focused on creating foundations instead of repurposing them should be more stimulated.

In this work, we seek to aid in improving some of these critical points and participate in the open development of some trends seen thus far. In the following sections, we present novel tools, datasets, and models for the Portuguese NLP community to expand upon. Although our efforts are mainly concerned with Brazilian Portuguese, we believe they can be repurposed, built upon, and adapted to other variants of Portuguese. In the following sections, we document the creation of our datasets, filtering methods, models, pretraining protocol, and evaluation procedures.

\section{Pretraining Data}
\label{section3}

\subsection{Concatenating GigaVerbo}
\label{section3-sub-1}

Datasets like the ones created by Lopes et al. \cite{lopes2024gl} (35.5B tokens) and Garcia et al. \cite{garcia2024robertalexpt} ($\approx$ 90B tokens) are filtered concatenations of several datasets used in previous studies or made accessible by crawling initiatives like Common Crawl and Oscar, much like the Pile \cite{gao2020pile}, and MassiveText \cite{rae2021scaling}, which are also collections of large text datasets from multiple sources, but with a focus on English. We applied the same methodology to create our dataset's initial version, concatenating several portions of openly available datasets for Portuguese and deduplicating their summation with an exact hash deduplication filter \cite{chenghao_mou_2023_8364980}. 

Our pretraining corpus, which we will refer to as \textit{GigaVerbo}, contains over 145 million documents, amounting to 780 GB of text. More details of its composition can be found in Table \ref{tab:gigaverbo}.

\begin{table}[h!]
    \centering
    \begin{tabular}{c | c | c | p{0.5\linewidth}}

    \textbf{Subset} & \textbf{Nº of Samples} & \textbf{\%} & \textbf{Description} \\
    \midrule
    
    \textbf{monoHPLT-PT}  & 58,244,012   & 40.09\%    & The clean and deduplicated Portuguese portion of the High-Performance Language Technologies resources dataset \cite{de2024new}. \\[2mm] 
    
    \textbf{CrawlPT}      & 43,846,974   & 30.17\%    & A deduplicated Portuguese corpus extracted from various web pages, concatenated from CC-100, Oscar, and BrWaC \cite{garcia2024robertalexpt, conneau2020unsupervised, OrtizSuarezSagotRomary2019, ortiz-suarez-etal-2020-monolingual, abadji2022towards, wagner2018brwac}. \\[2mm] 
    
    \textbf{Multilingual-C4} & 16,092,571  & 11.07\%    & The Brazilian Portuguese cleaned portion of the m-C4 dataset \cite{raffel2020exploring}. \\[2mm] 
    
    \textbf{Common Crawl} & 12,470,998   & 8.58\%     & A clean and deduplicated snapshot of the Common Crawl dataset (CC-MAIN-2023-23) \cite{wenzek-etal-2020-ccnet, conneau-etal-2020-unsupervised}. \\[2mm] 
    
    \textbf{BlogSet-BR}   & 4,321,181    & 2.97\%     & A collection of blog posts written in Brazilian Portuguese \cite{santos2018blogset}. \\[2mm] 
    
    \textbf{Instruct-PTBR} & 2,962,856   & 2.04\%     & A mix of multiple instruction datasets for various tasks, machine-translated from English to Brazilian Portuguese \cite{moro2024dataset}. \\[2mm] 
    
    \textbf{Corpus Carolina} & 2,075,395  & 1.43\%     & An open corpus with varied typology in contemporary Brazilian Portuguese \cite{corpusCarolinaV1}. \\[2mm] 
    
    \textbf{UltrachatBR}  & 1,255,091    & 0.86\%     & A Portuguese version (machine-translated) of the Ultrachat dataset \cite{ultrachatBr}. \\[2mm] 
    
    \textbf{Wikipedia}    & 1,101,475    & 0.76\%     & Cleaned Portuguese articles built from the Wikipedia dumps \cite{wikidump}. \\[2mm] 
    
    \textbf{CulturaX}     & 999,994      & 0.69\%     & The Portuguese portion of CulturaX, a multilingual dataset with 167 languages \cite{nguyen2023culturax}. \\[2mm] 
    
    \textbf{LegalPT}      & 925,522      & 0.64\%     & A concatenation of publicly available legal data in Portuguese, including legislation, jurisprudence, and legal articles \cite{sousa2019iudicium, bonifacio2020study, niklaus2023multilegalpile, rodrigues2023advancing, garcia2024robertalexpt}.\\[2mm] 
    
    \textbf{Gpt4All}      & 808,803      & 0.56\%     & A Portuguese (machine-translated) version of the Gpt4All dataset \cite{anand2023gpt4all}. \\[2mm] 
    
    \textbf{Bactrian-X}   & 66,994       & < 0.1\%    & The Portuguese portion of Bactrian-X, a collection of instruction-response pairs in 52 languages \cite{li2023bactrianx}.\\[2mm] 
    
    \textbf{XL-Sum}       & 64,577       & < 0.1\%    & A Portuguese (machine-translated) version of XL-Sum, a diverse dataset for abstractive summarization \cite{hasan-etal-2021-xl}.\\[2mm] 
    
    \textbf{Dolly 15K}    & 28,401       & < 0.1\%    & A Portuguese (machine-translated) version of Dolly 15K, an open-source dataset of instruction-following records generated by human annotators \cite{DatabricksBlog2023DollyV2}.\\[2mm] 
    
    \textbf{CosmosQA}     & 25,260       & < 0.1\%    & A Portuguese (machine-translated) version of the CosmosQA dataset for commonsense-based reading comprehension \cite{huang-etal-2019-cosmos}. \\[2mm] 
    
    \textbf{ROOTS}        & 10,740       & < 0.1\%    & The Portuguese portion of the ROOTS corpus, a dataset spanning 59 languages \cite{laurenccon2022bigscience}. \\[2mm] 
    
    \bottomrule
    \end{tabular}
    \vspace{0.25cm}
    \caption{Description of the different datasets comprising GigaVerbo. GigaVerbo is currently hosted on \href{https://huggingface.co/datasets/TucanoBR/GigaVerbo}{Hugging Face}. More information can be found in its dataset card.}
    \label{tab:gigaverbo}
\end{table}

\subsection{Filtering GigaVerbo}
\label{section3-sub-2}

As recent studies have suggested, several gains in performance can be achieved by enhancing dataset quality instead of merely scaling data ingestion and model size \cite{nguyen2022quality, gunasekar2023textbooks, li2023textbooks, penedo2024finewebdatasetsdecantingweb, wang2024finetuned, li2024datacomplmsearchgenerationtraining, tan20241, dubey2024llama}. However, what defines a text as "high-quality" is a nontrivial question. While heuristic-based filters can help us parse samples that are, for example, too short or ill-formatted, it is hard to differentiate high-quality text (e.g., articles, poems, tutorials) from plain text scrapped from the web (e.g., product information scrapped from e-commerce platforms) using only heuristic-based filters. Given that human annotation can be tedious and expensive \cite{dubois2024alpacafarm}, and current learned filters are either ill-suited for Portuguese or too expansive to run at scale, we decided to employ the same strategy used by Gunasekar et al. \cite{gunasekar2023textbooks} and train our own filtering system.

\begin{figure}
    \centering
    \includegraphics[width=400pt]{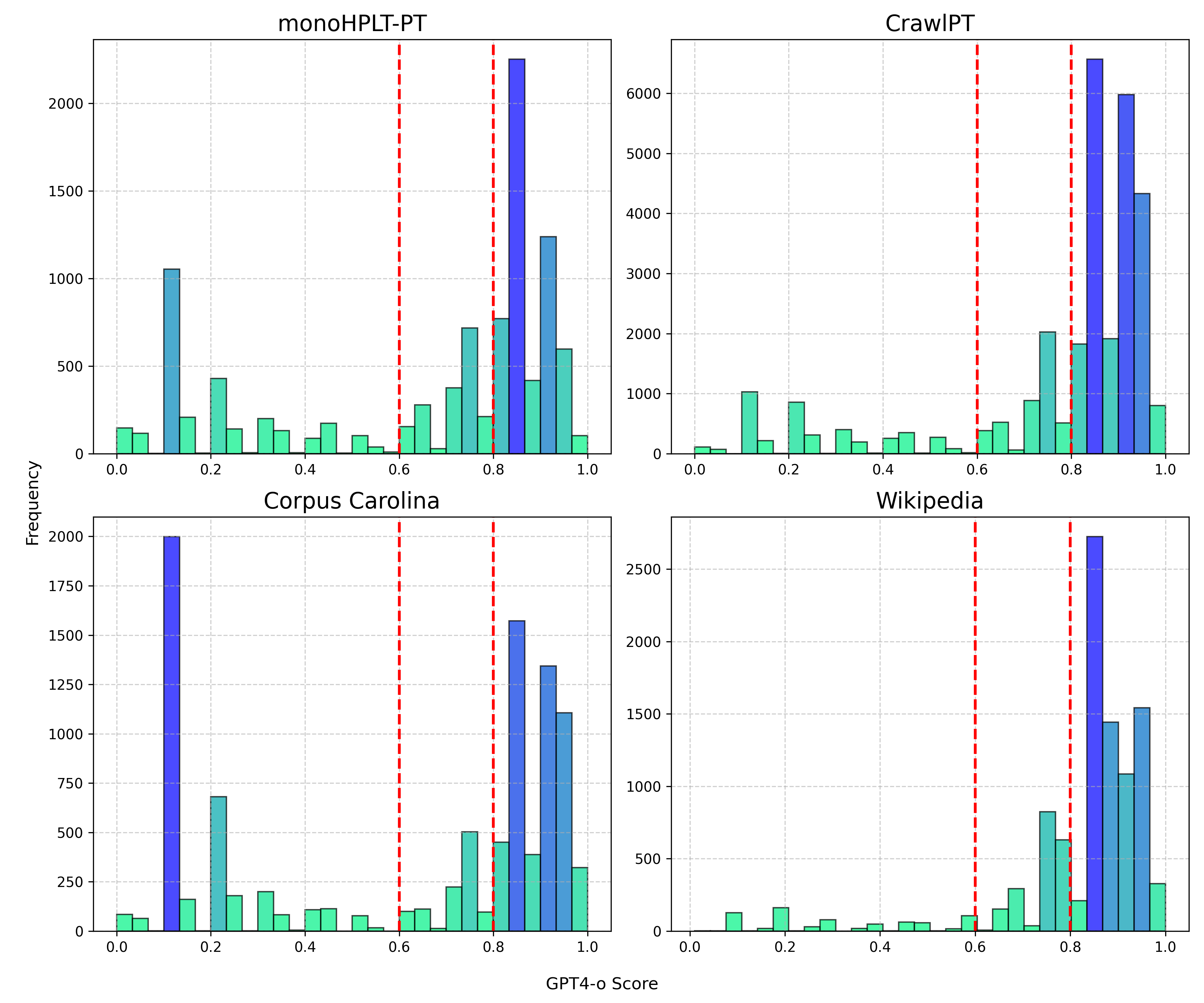}
    \caption{This graph shows the distribution of scores for 4 Subsets of GigaVerbo. We determined that the text  would have a "high" quality if the GPT-4o scores were >= 0.8 and "low" when <= 0.6, thus keeping our dataset with a more balanced proportion of labels for our classifiers. Above, we see that datasets like monoHPLT and Corpus Carolina have some of the lowest-quality samples. Also, given that GPT-4o is extremely sensitive to toxic and harmful content, samples containing toxic, dangerous, or NSFW content end up being scored very low (< 0.1), given as a way to account for the toxicity in our dataset. Analyzing samples from the Wikipedia portion scored by GPT-4o, we found that the model consistently gives low scores (< 0.5) to ill-formatted, incomplete, or excessively short documents (< 20 words). This classification/regression dataset is available on \href{https://huggingface.co/datasets/TucanoBR/GigaVerbo-Text-Filter}{Hugging Face}.}
    \label{fig:classdataset}
\end{figure}

For this, we randomly selected 110,000 samples from 9 Subsets of GigaVerbo (i.e., specifically those not synthetic).\footnote{These Subsets are monoHPLT-PT, CrawlPT, Wikipedia, CulturaX, Common Crawl, ROOTS, XL-Sum, Corpus Carolina, and LegalPT.} With these samples, we created a text-quality dataset using GPT-4o as a judge. Similar to the study of Gunasekar et al., \cite{gunasekar2023textbooks}, we prompted GPT-4o to score every text sample regarding its quality to create a high-quality text dataset for the Portuguese language (Fig. \ref{fig:classdataset}).\footnote{Example system prompt (translated): "\textit{You are required to act as a text classifier. Rate the quality of the text provided with a score between 0.0 and 1.0, considering how reasonable, valuable, and informative this text is for training a language model in Portuguese. Return the score to two decimal places without further comments}".}

As a first attempt, we sought to emulate Gunasekar et al. \cite{gunasekar2023textbooks} by converting the text samples of our classification dataset into embedding representations via a sentence-BERT \cite{reimers2019sentence}. After evaluating several available multilingual sBERTs, we selected LaBSE (Language-agnostic BERT Sentence Embedding) \cite{feng2020language}, which generates 768-dimensional embedding vectors. Then, we trained a shallow classifier based on XgBoost \cite{chen2016xgboost}. To convert Real numbered scores into labels, we binarized our data by defining as "high" all samples with a score >= 0.8 and "low" all those with a score <= 0.6. However, we were not satisfied with the results of this initial approach, and we hypothesize that the embedding representations of LaBSE were not performant enough for Portuguese. Hence, we decided to use BERTimbau \cite{souza2020bertimbau} as the foundation for a text classification model. Results for both approaches can be found in Table \ref{table:classperformance}.

\begin{table}[h!]
\centering
\begin{tabular}{lcccc}
\toprule
 & \textbf{Class} & \textbf{Precision} & \textbf{Recall} & \textbf{F1-score}\\
\midrule   

LaBSE + XGBoost & Low & 0.89 & 0.81 & 0.85 \\

 & High & 0.92 & 0.96 & 0.94 \\
\midrule   

BERTimbau & Low & 0.99 & 0.97 & 0.98 \\

& High & 0.99 & 0.99 & 0.99\\

\bottomrule
\end{tabular}
\vspace{0.25cm}
\caption{The table above shows the evaluation scores for both our LaBSE + XGBoost and BERTimbau-based classifiers. These scores were obtained by evaluating both models on a test set of 11,000 samples. For the XGBoost, we used a learning rate of 0.1, a maximum tree depth of 10, and 100 estimators. For fine-tuning BERTimbau, we used a learning rate of $4\times10^{-5}$, a weight decay of 0.01 for regularization, and a batch size of 128 for 3 epochs on our entire dataset. We also experimented with training a LaBSE + XGBoost regression algorithm, which achieved a root mean squared error of 0.16 on our evaluation, and the fine-tuning of BERTimbau-large, which achieved very similar results to its base version. All these models are available on \href{https://huggingface.co/TucanoBR}{Hugging Face}.}
\label{table:classperformance}
\end{table}

In the end, we chose to use our fine-tuned version of BERTimbau-base to filter GigaVerbo, given that it had achieved good performance and was faster than both our XGBoost classifiers and BERTimbau-large. After parsing GigaVerbo with our learned filter, from 145 million samples, our classifier assigned low-quality to approximately 50 million samples, leaving 65\% of GigaVerbo with a high-quality ranking according to our filter. However, for this study, we adopted a filtering approach where we only removed the low-quality samples if the confidence of our classifier was above 95\% for the low-quality class. We expect that this would minimize token waste due to low-confidence false negatives. This approach leaves us with $\approx$ 70\% of GigaVerbo to work with. The available GigaVerbo version on \href{https://huggingface.co/datasets/TucanoBR/GigaVerbo}{Hugging Face} has the class and confidence score assigned by our filter for each text sample, allowing other users to replicate our training mixture or adapt the filtering process to their liking.

\subsection{Scaling GigaVerbo}
\label{section3-sub-3}

According to the work of Muennighoff et al. \cite{muennighoff2023scaling}, training with up to 4 epochs of repeated data in data-constrained scenarios yields minor changes to loss compared to unique data, while further repetition yields less performance, eventually (for > 10 epochs) decaying to zero. Hence, to enlarge our pretraining corpus, when training the smaller versions of our series (i.e., 160m, 630m, and 1b1), we repeated specific GigaVerbo subsets based on the overall quality assigned by our learned filter. The contents of one epoch of GigaVerbo are shown in Table \ref{tab:scaled-gigaverbo}. To train our largest model (2b4), we repeated the entire filtered dataset for four epochs.

\begin{table}[htbp]
    \centering
    \begin{tabular}{l|r|r|r|r|r}
    
    \toprule
    \textbf{Subset}          & \textbf{Original Size} & \textbf{Filtered Size} & \textbf{\%} & \textbf{Repeat Factor} & \textbf{Token Count} \\
    \midrule   

    monoHPLT-PT    & 58,244,012   & 37,291,607  & 64.03\%   & 1 & 84,708,988,928   \\[2mm] 
    CrawlPT        & 43,846,974   & 29,427,715  & 67.11\%   & 1 & 14,023,256,064   \\[2mm] 
    Multilingual-C4 & 16,092,571   & 13,849,412  & 87.10\%   & 2 & 8,083,937,280   \\[2mm] 
    Common Crawl   & 12,470,998   & 10,527,584  & 84.42\%   & 2 & 14,421,852,160   \\[2mm] 
    BlogSet-BR     & 4,321,181    & 2,411,590   & 55.81\%   & 1 & 1,561,569,280    \\[2mm] 
    Instruct-PTBR  & 2,962,856    & 2,570,829   & 86.77\%   & 4 & 1,141,768,192    \\[2mm]
    Corpus Carolina & 2,075,395    & 1,170,905   & 56.42\%   & 1 & 1,018,951,680   \\[2mm]
    UltrachatBR    & 1,255,091    & 1,2477,14   & 99.41\%   & 4 & 1,652,916,224    \\[2mm]
    Wikipedia      & 1,101,475    & 921,137     & 83.63\%   & 4 & 551,403,520      \\[2mm]
    CulturaX       & 999,994      & 883,550     & 88.36\%   & 4 & 565,768,192      \\[2mm] 
    LegalPT        & 925,522      & 891,891     & 97.62\%   & 4 & 1,313,269,760    \\[2mm] 
    Gpt4All        & 808,803      & 725,195     & 89.66\%   & 4 & 381,650,944      \\[2mm]
    Bactrian-X     & 66,994       & 55,685      & 83.012\%   & 4 & 9,517,056       \\[2mm]
    XL-SUM         & 64,577       & 64,467      & 99.83\%   & 4 & 52,072,448       \\[2mm]
    Dolly 15K      & 28,401       & 21,016      & 74.00\%   & 2 & 3,698,688        \\[2mm] 
    CosmosQA       & 25,260       & 14,702      & 58.20\%   & 1 & 2,074,624        \\[2mm]
    ROOTS          & 10,740       & 5,448       & 50.72\%   & 1 & 11,456,512       \\[2mm]

    \textbf{Total} & \textbf{145,300,844} & \textbf{102,080,447} & \textbf{70.25\%} &   & \textbf{129 Billion} \\ 
    \bottomrule
    \end{tabular}
    \vspace{0.25cm}
    \caption{In the table above, we present the number of documents present in every subset of GigaVerbo (i.e., its original size), its size after filtering, and the repetition factor used for creating the data mixture used to train the 160m, 630m, and 1b1 versions of the Tucano series, which generates a dataset with 169 billion tokens. The token count column provides raw values, i.e., the token count without accounting for the repetition factor of the filtered portion of GigaVerbo (129 billion tokens). Without filtering, GigaVerbo contains $\approx$ 200 billion tokens. To train our biggest model (Tucano-2b4), we repeated the entire filtered dataset for four epochs, amounting to $\approx$ 515 billion tokens.}
    \label{tab:scaled-gigaverbo}
\end{table}



\section{Tokenization}
\label{section4}

As already pointed out by previous studies \cite{finardi2021berta, cui2023efficient, larcher2023cabrita, correa2024teenytinyllama}, the success of a tokenization scheme in compressing a given language has a subsequent impact on the efficiency of the language model in question. While the precise effect on the overall language modeling capability remains unclear \cite{schmidt2024tokenization}, the tokenization scheme certainly plays a significant role in this process \cite{goldman2024unpacking}. In terms of compression, one can significantly improve tokenizer efficiency (i.e., how many tokens are required to encode a given piece of text) when using a vocabulary custom-made for a given domain \cite{larcher2023cabrita, correa2024teenytinyllama}. This allows us to better utilize limited resources, like context, when working with transformer-based models. 

\begin{figure}
    \centering
    \includegraphics[width=400pt]{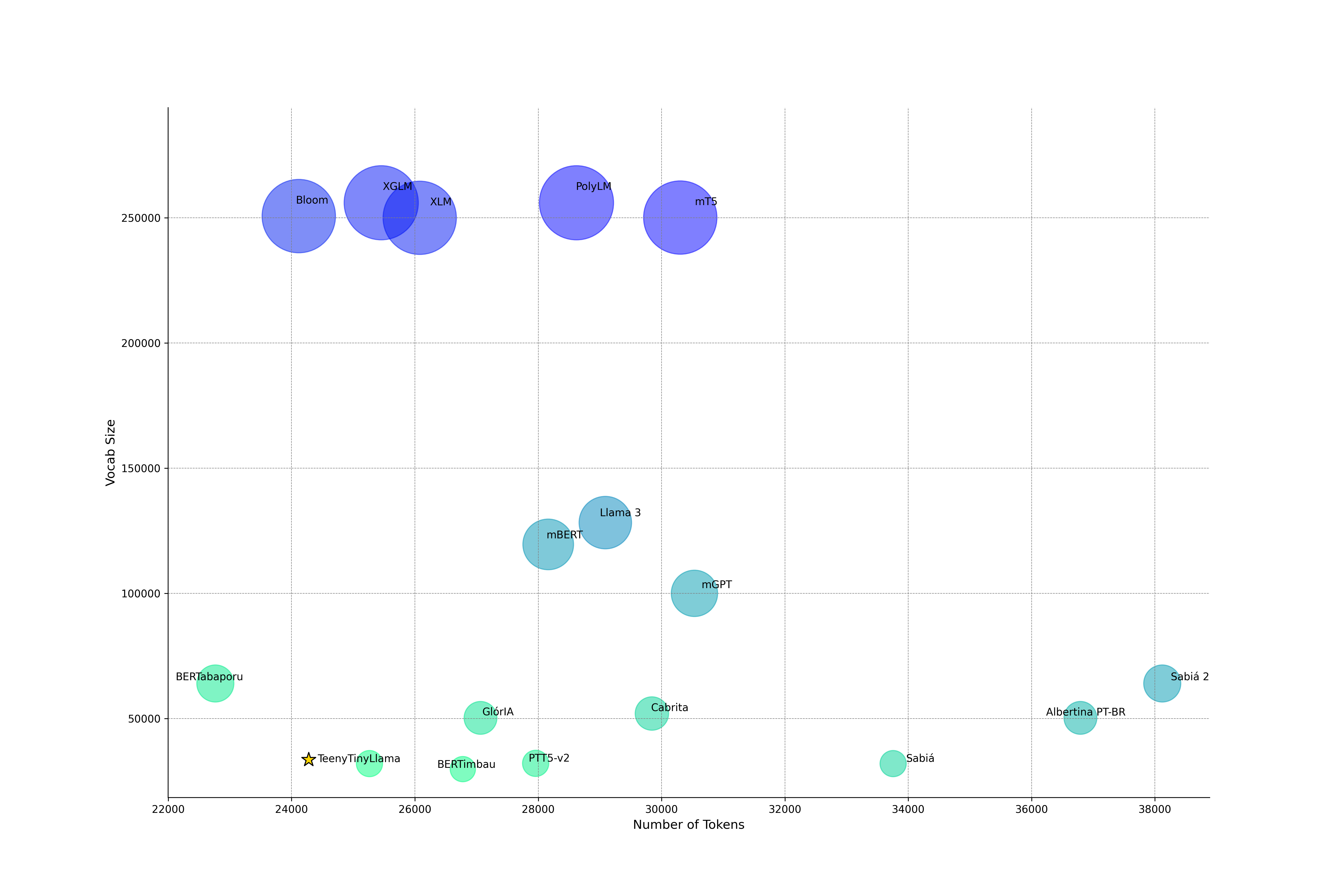}
    \caption{The figure above lets us understand specific relationships between vocabulary size and the respective tokenizer's capabilities regarding compression. For example, models that use the Llama 2 tokenizer (e.g., Sabiá), primarily focused on English, do not encode Portuguese very efficiently. On a similar note, Sabiá-2 has the worst performance across all tokenizers, even though it has double the vocab size of its predecessor. Meanwhile, multilingual models, like mBERT, PolyLM, Llama 3, mT5, and mGPT, improve their compression efficiency by having significantly enlarged vocabularies, with Bloom, XGLM, and XLM being close to the top of this comparison, all using massive multilingual vocabularies with > 250,000 tokens. As a middle ground between efficiency and resource consumption (i.e., larger vocabularies imply larger embedding matrices, which then imply more computational requirements for inference or training), we have tokenizers with vocabularies tailored for the Portuguese domain (e.g., BERTabaporu, TeenyTinyLlama, BERTimbau). In summary, while multilingual (or larger) vocabularies generally offer improved compression, small, domain-specific tokenizers balance efficiency and computational resource consumption. The code for replicating this test is available in \href{https://github.com/Nkluge-correa/Tucano/tree/main/logs/README.md}{GitHub}.}
    \label{fig:tokenizer}
\end{figure}

To better assess and compare tokenizer efficiency across our revised anthology of Portuguese language models, we replicated the test evaluation performed by both Larcher et al. \cite{larcher2023cabrita} and Corrêa et al. \cite{correa2024teenytinyllama} on several available tokenizers tied to Portuguese LLMs. For this, we used a text sample containing $\approx$ 14,000 words from Portuguese poems extracted from authors like Fernando Pessoa and Ronald de Carvalho, among others. Our results are displayed in Fig. \ref{fig:tokenizer}.

According to our experiments, the tokenizer trained by Corrêa et al. \cite{correa2024teenytinyllama} presents both an efficient compression capability and a slim vocabulary size for improved efficiency during input and output embedding matrices computations. The TeenyTinyLlama tokenizer (from now on referred to as the \textit{Tucano} tokenizer) is a Sentencepiece tokenizer \cite{kudo2018sentencepiece}, which implements both sub-word and unigram tokenization. Finally, we utilized this tokenizer to encode our pretraining dataset, separating each document with an end-of-text token (\texttt{</s>}).

\section{Architecture}
\label{section5}

Like many other studies \cite{pires2023sabi, larcher2023cabrita, correa2024teenytinyllama, bode2024, almeida2024sabi}, we used a decoder-only Transformer based on the Llama architecture \cite{touvron2023llama1, touvron2023llama2, dubey2024llama} as the basis for our models. In terms of code implementation, we used the implementation provided by Hugging Face so that our models can be easily shared and used by the community. Like the standard Llama architecture, our models use both root mean square layer normalization \cite{zhang2019root} and RoPE embeddings \cite{su2021roformer}, with Silu activation's \cite{elfwing2018sigmoid} instead of the SwiGLU \cite{shazeer2020glu} described in the original Llama papers. All models were trained using a causal language modeling objective and cross-entropy as its loss. The dimensions of our models, which we named \textit{Tucano}, are documented in Table \ref{tab:model}.

\begin{table}[htbp]
  \centering
  \scalebox{1.0}{
  \begin{tabular}{lcccccccc}
    \toprule
    $n_{param}$ & $n_{layers}$ & $d_{model}$ & $d_{mlp}$ & $n_{heads}$ & $n_{KV-heads}$ & $d_{head}$ & $c_{length}$ \\ 
    
    \midrule
    
    162,417,408 & 12 & 768 & 3,072 & 12 & 12 & 64 & 2048 \\
    630,253,568 & 14 & 2,048 & 4,096 & 16 & 4 & 128 & 2048  \\
    1,100,048,384 & 22 & 2,048 & 5,632 & 32 & 4 & 64 & 2048\\
    2,444,618,240 & 24 & 2,560 & 10,240 & 16 & 4 & 160 & 4096 \\

    \bottomrule
  \end{tabular}
  }
  \vspace{0.25cm}
  \caption{Each model is based on a decoder-only Llama architecture, with a vocabulary size of 32,000. Tucano-160m, 630m, and 1b1 were trained with a context window of 2048 tokens, while the largest model (2b4) was trained with sequences of length 4096. All models were trained using a causal language modeling objective and cross-entropy loss. The parameters were explicitly tuned to fit these models (and respective optimizers and batches) on A100-SXM4-80GB GPUs. Unlike previous studies \cite{correa2024teenytinyllama, dey2023cerebras}, we trained all models beyond what the Chinchilla scaling laws prescribed \cite{hoffmann2022training}. Group-query attention \cite{ainslie2023gqa}, with 4 key-value heads per decoder block, was used to reduce attention computations' memory footprint, helping us to maximize token throughput during training without significantly impacting model convergence when training Tucano-630m, 1b1, and 2b4.}
  \label{tab:model}
\end{table}

\section{Training and Evaluation}
\label{section6}

\subsection{Hyperparameters and Performance}
\label{section6-sub-1}

Our training code base was primarily built with a PyTorch-based deep learning stack \cite{paszke2019pytorch}. As a training framework, given that our model sizes could all fit inside the memory of our GPUs, we utilized a simple Distributed Data-Parallel approach \cite{li2020pytorch}. For support, we used specific libraries tied to the Hugging Face ecosystem, like Tokenizers \cite{tokenizers} for fast tokenization and Datasets \cite{lhoest-etal-2021-datasets} for handling our pretraining corpus. We also used FlashAttention \cite{dao2022flashattention, dao2023flashattention2} for optimized IO-aware attention computation and the Liger Triton kernels \cite{liger2024} to reduce memory footprint and improve token throughput during training. We used 8 NVIDIA A100-SXM4-80GB GPUs to train both smaller versions of Tucano (160m and 630m) and 16 of these for our two largest models (1b1 and 2b4). Lastly, we utilized CodeCarbon \cite{codecarbon} to track the resource consumption of our experiments and training runs.

To assess the efficiency of our deep learning stack, we utilized the method proposed by Chowdhery et al. \cite{chowdhery2022palm} to estimate the model FLOPs utilization (MFU) we were able to achieve during our training runs, which can be understood as the ratio of the observed throughput (actual performance) to the theoretical maximum throughput that a given hardware offers. Regarding speed comparison, our code implementation is on par with other documented developments in the literature. For example, for our Tucano-1b1, we were able to achieve a training throughput of 24,180 tokens per second per A100-SXM4-80GB, which is similar to that achieved in the development of TinyLlama \cite{zhang2024tinyllama}, and superior to those documented in the development of the Pythia suite \cite{biderman2023pythia}. In Table \ref{tab:training-config}, we document the hyper-settings used to train our models and the efficiency we achieved.

\begin{table}[htbp]
  \centering
  \scalebox{1.0}{
  \begin{tabular}{p{0.4\linewidth} | ccccc}
    \toprule
     & 160m & 630m & 1b1 & 2b4 \\
    \midrule
    
    Total Optimization Steps & 320K & 400K & 480K & 1.9M \\
    Batch Size (in tokens) & 524K & 524K & 524K & 262K \\
    Epochs & 1 & 1.25 & 1.5 & 4 \\
    Total Tokens & 169B & 211B & 250B & 515B \\
    Total Time (days) & 1.8 & 6.9 & 7.2 & 30 \\
    Tokens/parameter & 1,050 & 335 & 225 & 210 \\
    Max Learning Rate & $1\times10^{-3}$ & $6\times10^{-4}$ & $4\times10^{-4}$ & $2\times10^{-4}$\\
    Min Learning Rate & $1\times10^{-4}$ & $6\times10^{-5}$ & $4\times10^{-5}$ & $2\times10^{-5}$ \\
    GPU Count (A100) & 8 & 8 & 16 & 16 \\
    MFU & 43\% & 54\% & 53\% & 55\% \\
    Tokens/seconds & 1,066,000 & 346,000 & 387,000 & 180,200 \\
    \% Memory Footprint & 43.75\% & 92.5\% & 95\% & 95\%\\

    \bottomrule
  \end{tabular}
  }
  \vspace{0.25cm}
  \caption{All models used AdamW \cite{loshchilov2017decoupled}, with the following configuration: $\varepsilon$ = $1\times10^{-8}$, $\beta_{1}$ = 0.9, and $\beta_{2}$ = 0.95. We applied a weight decay factor of 0.1 and a gradient clipping threshold of 1.0 to regularize gradient values. Regarding optimizer scheduling, all models had 1,000 warm-up steps, where the learning rate was linearly increased up to the max learning rate. After that, during 90\% of the training, we used a cosine learning rate decay from its maximum value to a minimum learning rate (10\% of the maximum learning rate). For the last 10\% of the training, the minimum learning rate is sustained as a constant. All models were trained using BF16 mixed precision, TF32 mode enabled for matrix multiplication operations, and FlashAttention 2, in addition to the Liger Triton kernels. Many of these configurations were estimated via experiments (i.e., short runs of $\approx$ 10,000 steps) or directly imported from the documentation of other LLMs of similar size \cite{brown2020language, zhang2022opt, workshop2022bloom, biderman2023pythia, gunasekar2023textbooks, li2023textbooks, zhang2024tinyllama, tan20241}.}
  \label{tab:training-config}
\end{table}

\subsection{Batch Size and Gradient Accumulation}
\label{section6-sub-2}

According to the literature, transformer-based networks can benefit from larger batch sizes during training \cite{shallue2019measuring}. By larger, we mean up to millions of tokens per batch. For example, in the first iteration of GPT-3, the series was trained on batches from 524K to 3.2 million tokens \cite{brown2020language}, with batch sizes increasing with model size. Meanwhile, all Llama 2 models were trained with 4 million tokens per batch \cite{touvron2023llama2}, while Llama 3 405B used a massive amount of 16 million tokens per batch \cite{dubey2024llama}. In Biderman et al. \cite{biderman2023pythia} development of the Pythia suite, all models were trained with a batch size of 2 million. Currently, for language model training at the sub 2 billion parameters mark, most studies maintain a batch size between 1 to 2 million tokens per batch \cite{gunasekar2023textbooks, zhang2024tinyllama, lopes2024gl, tan20241}.

Given that achieving this batch size range requires that our hardware have a significant amount of available memory for batch processing, a common approach documented in the literature is using \textit{gradient accumulation} strategies when limited by available VRAM. In essence, gradient accumulation is used during training to simulate larger batch sizes than our hardware's memory typically allows. In this approach, instead of updating the model parameters after each mini-batch, the gradients are computed and stored in several gradient accumulation steps. Still, the model weights are not updated immediately. Instead, the gradients are accumulated and normalized over multiple mini-batches, and only after a specified number of iterations (the accumulation steps) are the model's parameters updated. This method effectively allows training with larger equivalent batch sizes without increasing memory requirements. Hence, several studies document the use of this approach to simulate large, million-size batch ranges \cite{gunasekar2023textbooks, larcher2023cabrita, lopes2024gl}. 

\begin{figure}
    \centering
    \includegraphics[width=400pt]{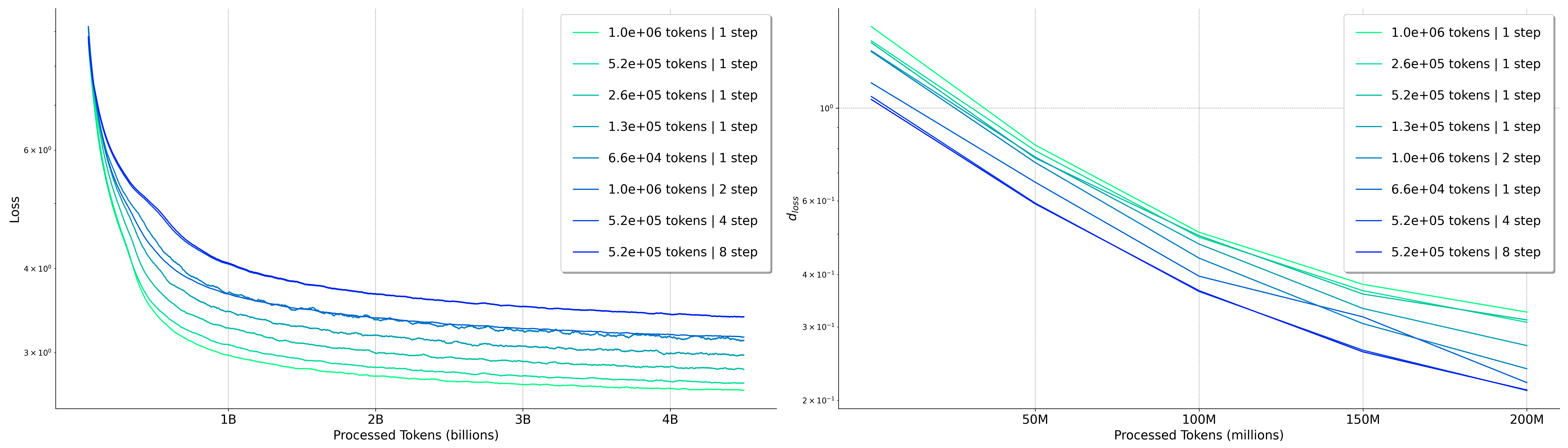}
    \caption{We tested several batch sizes on our 160 million parameter model, from 1 million (512) to 65 thousand tokens (32), while also trying to reproduce a 0.5 million batch (256) via different levels of gradient accumulation (i.e., 2, 4, and 8 accumulation steps). We maintained the learning rate and the $\beta$ values of the AdamW constant for all these tests, together with a linear warm-up of 1,000 steps. As expected, the 1 million tokens batch, with no gradient accumulation steps (i.e., step of 1), produced the best loss curve with faster convergence at the earliest stages of training, followed by all other batch sizes (i.e., 256, 128, 64, and 32) that did not have gradient accumulation steps. At the same time, the more gradient steps are applied to achieve a desired batch size, the slower the convergence rate, up to the point that training with a global batch size of 32 and a global batch of 512 achieved via 2 gradient accumulation steps yield the same results in terms of convergence speed. While the plot on the left shows the shape of the loss curve for several different batch sizes and gradient accumulation configurations, the plot on the right shows the rate of change in loss ($d_{loss}$) for the first 200 million tokens. While this rate of change tends to converge to the same value for all experimented batch sizes (i.e., with time, all lines converge at the same rate), the initial values differ significantly in the early stages of training, with bigger "natural" batches presenting a higher rate of change. Although not through extensive exploration, we observed the same behavior for our other model sizes, independent of tweaks to the learning rate hyper-settings or changes in the number of warm-up steps.}
    \label{fig:ga-experiments}
\end{figure}

Aware of this trend, our initial experiments used gradient accumulation steps to increase our models' batch size artificially. However, we documented a significant decrease in convergence speed when applying gradient accumulation steps, where the more gradient accumulation steps performed (e.g., 2, 4), the slower the convergence of our models became. To investigate this issue further, we promoted a series of small test runs on our smallest model to track how the rate of change in loss ($d_{loss}$) was influenced by the amount of gradient accumulation steps performed. Our results are depicted in Fig. \ref{fig:ga-experiments}.

Given these results, we conclude that a better performance can be achieved \textit{without the use any form of gradient accumulation}. Although we could have lowered our memory footprint by using gradient checkpointing \cite{chen2016training}, and hence, increased our batches without the need for accumulation steps, we decided to prioritize training speed and efficient hardware utilization. It is puzzling why other works that employed the gradient accumulation approach on Llama models did not report such a phenomenon \cite{larcher2023cabrita}. For the moment, we hypothesize that the root means squared normalization used in the Llama architecture, which already averages batch-dependent activations, suffered from an introduction of extra noise when we adopted smaller batches combined with gradient accumulation averaging of the loss. However, more extensive experimentation is required to truly understand the causes of this phenomenon and determine how practitioners can use gradient accumulation strategies without severely impacting the training performance of this type of neural architecture.

\subsection{Evaluation Protocol}
\label{section6-sub-3}

While training our base models, we saved several checkpoints for each model at intervals of approximately 10.5 billion tokens. For every checkpoint, in addition to running a small evaluation dataset (i.e., 60,000 samples randomly selected and excluded from GigaVerbo) to track the model's perplexity as training progressed, we employed a comprehensive evaluation harness. This harness was modeled after the work of Corrêa et al. \cite{correa2024teenytinyllama}, with additional benchmarks included. The benchmarks in our evaluation harness can generally be categorized into two types: \textit{native} evaluations, specifically developed in Portuguese, and \textit{translated} ones, consisting of English benchmarks machine-translated into Portuguese.

Although native benchmarks are ideal for assessing linguistic and cultural nuance, using translated datasets was necessary due to the limited availability of Portuguese-specific evaluation benchmarks. This approach allows us to determine our model's generalization capabilities across various domains and ensure that critical evaluation categories, often well-represented in English datasets, are dutifully assessed. Table \ref{tab:benchmarks} lists all the evaluations used in our custom harness.

\begin{table*}[h]
    \centering
    \begin{tabular}{p{0.3\linewidth} | cccc}
    
    \toprule
    \textbf{Benchmark} & \textbf{n-shot} & \textbf{Origin} & \textbf{Type} & \textbf{Metric} \\
    \midrule
    
    ENEM & 3-shot & Native & Q\&A & \texttt{acc} \\[2mm]
    BLUEX & 3-shot & Native & Q\&A & \texttt{acc} \\[2mm]
    OAB Exams & 3-shot & Native & Q\&A & \texttt{acc} \\[2mm]
    ASSIN2 RTE & 15-shot & Native & Entailment & \texttt{f1 macro} \\[2mm]
    ASSIN2 STS & 10-shot & Native & Similarity & \texttt{pearson} \\[2mm]
    FAQUAD NLI & 15-shot & Native & Entailment & \texttt{f1 macro} \\[2mm]
    HateBR & 25-shot & Native & Classification & \texttt{f1 macro} \\[2mm]
    PT Hate Speech & 25-shot & Native & Classification & \texttt{f1 macro} \\[2mm]
    TweetSentBR & 25-shot & Native & Classification & \texttt{f1 macro} \\[2mm]
    CALAME-PT & 0-shot & Native & Next Word Prediction & \texttt{acc} \\[2mm]
    ARC-Challenge & 25-shot & Translated & Q\&A & \texttt{acc norm} \\[2mm]
    HellaSwag & 10-shot & Translated & Q\&A & \texttt{acc norm} \\[2mm]
    TruthfulQA & 0-shot & Translated & Q\&A &\texttt{bleurt}\\[2mm]
    LAMBADA & 0-shot & Translated & Next Word Prediction & \texttt{acc} \\

    \bottomrule
    \end{tabular}
    \caption{A description of the evaluation harness used in our work. To learn how to replicate our usage of this harness, please visit the evaluation section of our \href{https://github.com/Nkluge-correa/Tucano/tree/main/evaluations/README.md}{GitHub} repository.}
    \label{tab:benchmarks}
\end{table*}

In total, this evaluation harness comprises 14 benchmarks, ten of which are native to Portuguese, and four are machine-translated from English datasets. The native benchmarks include ENEM \cite{ENEM-Challenge}, BLUEX \cite{almeida2023bluex}, OAB \cite{delfino2017passing}, ASSIN2 RTE \cite{real2020assin}, ASSIN2 STS \cite{real2020assin}, FAQUAD NLI \cite{faquad-nli-2029}, HateBR \cite{vargas2022hatebr}, PT Hate Speech \cite{fortuna2019hierarchically}, and TweetSentBR \cite{brum2017building}, all tied to a natively Brazilian Portuguese implementation of the Language Model Evaluation Harness \cite{gao2021framework}, made available by Garcia \cite{open-pt-llm-leaderboard}. In the assessment of CALAME-PT, we had to create a custom evaluation protocol based on the work of Lopes et al. \cite{lopes2024gl}, i.e., all generations are performed deterministically without sampling in a zero-shot manner, with only exact matches being counted as a successful inference.

The remaining four benchmarks, ARC-Challenge \cite{clark2018think}, HellaSwag \cite{zellers2019hellaswag}, and TruthfulQA \cite{lin2021truthfulqa} are all evaluations tied to a machine-translated version of the original (English) datasets, made available by a multilingual implementation of the Language Model Evaluation Harness (Lai et al. \cite{lai2023okapi}). All few-shot settings for assessment remain the same as the one set for standard leaderboard comparisons. For LAMBADA-PT, a machine-translated version of the original test set of LAMBADA \cite{paperno2016lambada}, we used the same evaluation protocol as the one used in CALAME-PT, given that both benchmarks involve the same primary task (i.e., predict the final word of a given sentence).

Finally, to evaluate the "Instruct" versions (see Section \ref{section7} for more details) of our base models, we developed a Portuguese chat evaluation dataset, comprised of 805 completion samples extracted from a machine-translated version of the original Alpaca dataset \cite{taori2023alpaca}\footnote{This dataset is available on \href{https://huggingface.co/datasets/TucanoBR/alpaca-eval-pt}{Hugging Face}.}. In this evaluation, our model's outputs are compared to a reference standard\footnote{In our case, we use the original \texttt{text-davinci-003} completions from the Alpaca dataset.} and later judged by an automated annotator (GPT-4 Turbo) to determine their relevance, coherence, and adherence to the instruction prompts. In line with the evaluation methodology proposed by Dubois et al. \cite{dubois2024length}, we use length-controlled win rates as our evaluation metric, given that these are highly correlated with human preferences and evaluations of pair-wise comparisons.

\section{Alignment}
\label{section7}

To offer a more easy-to-use version of our more capable models (i.e., 1b1 and 2b4), we implemented a fine-tuning process divided into two stages: supervised fine-tuning (SFT) \cite{ouyang2022training} and direct preference optimization (DPO) \cite{rafailov2024direct}.

For the supervised fine-tuning step, we synthesized a small dataset containing over 600K samples of user-assistant interactions generated by other models that went through an alignment process.\footnote{In general terms, we can define an alignment process as a process in which we seek to improve a system's capability to follow human instructions and intentions while minimizing the possible harm it can cause \cite{correa2024dynamic}.}. A description of this dataset can be found in Table \ref{tab:sft-dataset}. These fine-tuned models have special chat-delimiting tokens (i.e., \texttt{<instruction>} and \texttt{</instruction>}) added to their tokenizers and were trained by starting from the latest checkpoint of their respective model (e.g., Tucano-1b1, step 480,000). Regarding hyper-settings, fine-tuning jobs performed another learning rate decay to 10\% of the original minimal value achieved during training, with no warm-up steps and all other hyper-parameters unchanged. Both models were trained on a batch size of 262K tokens per optimizer step for four epochs.

\begin{table}[h!]
    \centering
    \begin{tabular}{c | c | c | p{0.45\linewidth}}

    \textbf{Subset} & \textbf{Nº of Samples} & \textbf{\%} & \textbf{Description} \\
    \midrule
    
    \textbf{GPT4-500k-PTBR}  & 565,536   & 83\%    & A machine-translated version of conversations with GPT-4 \cite{moro2024dataset2}. \\[2mm]

    \textbf{Orca-Math-PT}  & 64,073   & 9.5\%    & A machine-translated version of Orca-math dataset \cite{mitra2024orca}. \\[2mm] 

    \textbf{Instruct-Aira v.3}  & 50,000   & 7.5\%    & A collection of multi-turn conversations generated by user interactions with conversational LLMs \cite{correa2024dynamic}. \\[2mm]
    
    \bottomrule
    \end{tabular}
    \vspace{0.25cm}
    \caption{A description of the datasets used in the alignment of the trained models. This dataset is currently hosted on \href{https://huggingface.co/datasets/TucanoBR/Tucano-SFT}{Hugging Face}. More information can be found in its dataset card.}
    \label{tab:sft-dataset}
\end{table}

Finally, for the DPO step, we used the preference modeling dataset developed by Corrêa \cite{correa2024dynamic}, which consists of 35K triplets comprising a user prompt, a preferred response, and a less preferred alternative. We design our DPO fine-tuning implementation on top of the Transformer Reinforcement Learning (TRL) library \cite{vonwerra2022trl}. We trained both models using their respective SFT versions as initial checkpoints. Regarding hyper-settings, for both models, we used a cosine learning rate scheduler with a learning rate of $1\times10^{-6}$ and a weight decay of 0.1. We set \texttt{beta} to 0.5, applied sigmoid as the loss function and used zero label smoothing. We repeated the dataset for three epochs, with a global batch size of 16 for the 1b1 model and 8 for the 2b4 model. This two-step alignment approach outputs the "Instruct" version of our models: Tucano-1b1-Instruct and  Tucano-2b4-Instruct.

\section{Results}
\label{section8}

\begin{figure}[hbt!]
    \centering
    \includegraphics[width=400pt]{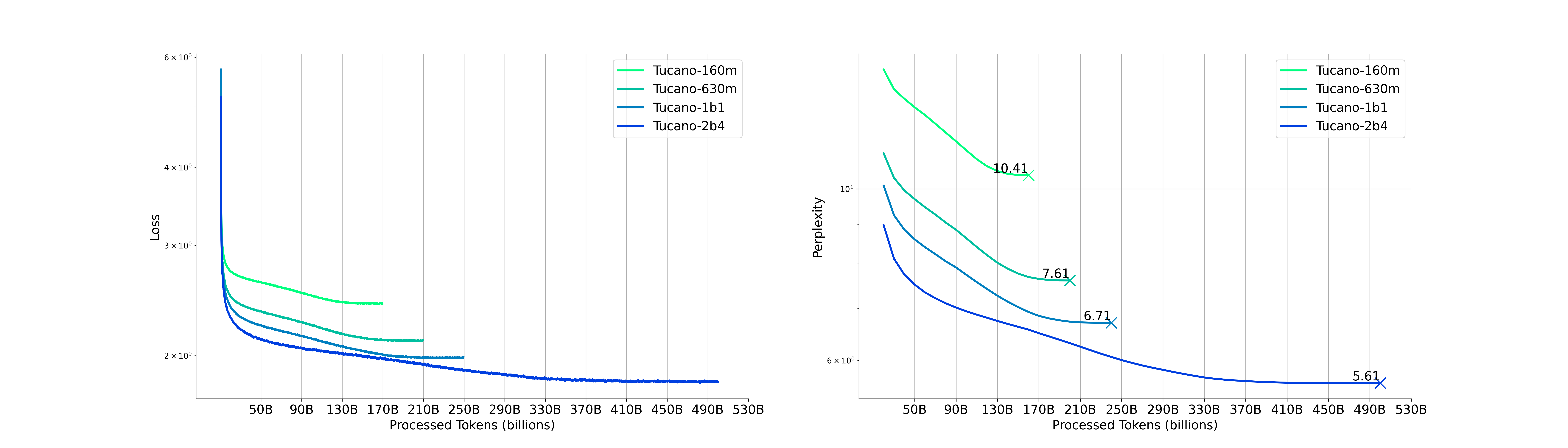}
    \caption{All logs from our training runs recorded loss, evaluation loss, the current value of the learning rate, and the gradient norm for that specific optimization step. These logs are available in our \href{https://github.com/Nkluge-correa/Tucano/tree/main/logs/README.md}{GitHub} repository.}
    \label{fig:loss-perplexity}
\end{figure}

Figure \ref{fig:loss-perplexity} depicts the logged training loss and validation perplexity curves for all four base models we trained. As expected, larger models exhibit a more significant reduction in loss and perplexity as training progresses, even though this difference would be made more pronounced if we could train our bigger models with larger batches. In short, our logs reaffirm that as the model size and data ingestion are increased, the performance of the language model also increases.

\subsection{Benchmark Evaluations}
\label{section8-sub-1}

As mentioned, for every 10.5 billion tokens processed during training, we saved a checkpoint for each model and ran our evaluation harness on it. This approach allowed us to systematically track and represent model performance as a function of time and token ingestion, enabling us to observe how model performance, across several benchmarks, is related to token ingestion on a plain causal language modeling regime without intentionally seeking to overfit a specific training (or evaluation) distribution. 

If we assume that \textit{"the more a model is trained on new tokens, the more capable it becomes"}, as demonstrated by several works examining the scaling behavior of LLMs \cite{rae2021scaling, hoffmann2022training, biderman2023pythia, xue2023repeat, zhang2024tinyllama}, we would expect to observe this phenomenon when evaluating our models with the custom evaluation harness we developed (which contains most of the evaluations used by the Portuguese NLP community). To test this hypothesis, we calculated the Pearson product-moment correlation coefficients between our evaluation results and the number of tokens processed at each checkpoint. A positive correlation between token ingestion and benchmark performance would suggest a relationship between these variables, implying that performance improves as the model ingests more tokens. However, this anticipated pattern was only observed across some benchmarks, as seen in Table \ref{tab:benchmark-corr}.

\begin{table*}[h]
    \centering
    \begin{tabular}{p{0.3\linewidth} | cccc}
    
    \toprule
    \textbf{Benchmark} & $r_{\text{160m}}$ & $r_{\text{630m}}$ & $r_{\text{1b1}}$ & $r_{\text{2b4}}$ \\
    \midrule
    
    ENEM & 0.10 & -0.45 & 0.41 & 0.48 \\[2mm]
    BLUEX & -0.06 & 0.21 & 0.52 & 0.00 \\[2mm]
    OAB Exams & 0.34 & 0.21 & 0.28 & 0.24 \\[2mm]
    ASSIN2 RTE & -0.30 & \textbf{0.78} & \textbf{0.74} & 0.34 \\[2mm]
    ASSIN2 STS & 0.58 & 0.22 & \textbf{0.81} & -0.50 \\[2mm]
    FAQUAD NLI & -0.31 & 0.00 & 0.00 & 0.17 \\[2mm]
    HateBR & -0.56 & \textbf{0.65} & 0.48 & -0.36 \\[2mm]
    PT Hate Speech & -0.75 & -0.15 & 0.31 & 0.49 \\[2mm]
    TweetSentBR & 0.17 & 0.50 & 0.33 & \textbf{0.88} \\[2mm]
    \textbf{CALAME-PT} & \textbf{0.92} & \textbf{0.96} & \textbf{0.94} & \textbf{0.86} \\[2mm]
    \textbf{ARC-Challenge} & \textbf{0.76} & \textbf{0.69} & \textbf{0.82} & \textbf{0.61} \\[2mm]
    \textbf{HellaSwag} & \textbf{0.93} & \textbf{0.94} & \textbf{0.90} & \textbf{0.89} \\[2mm]
    TruthfulQA & -0.16 & \textbf{0.66} & -0.05 & -0.30 \\[2mm]
    \textbf{LAMBADA} & \textbf{0.89} & \textbf{0.90} & \textbf{0.91} & \textbf{0.86} \\

    \bottomrule
    \end{tabular}
    \caption{The table shows all correlation scores for each benchmark against the different models. The highlighted scores correspond to a Pearson product-moment correlation value above 0.6, while the highlighted benchmarks are those where a positive correlation above 0.6 was found for all models, irrespective of size. To replicate these results, you can use the evaluation logs and code implementation available in our \href{https://github.com/Nkluge-correa/Tucano/tree/main/logs/README.md}{GitHub} repository.}
    \label{tab:benchmark-corr}
\end{table*}

Even when considering the possibility that specific in-context capabilities only \textit{emerge} once models reach a particular scale, our results do not consistently show this pattern across benchmarks of the same type, such as multiple-choice Q\&A evaluations. For instance, benchmarks like ENEM and BLUEX show a moderate positive correlation only for the 1b1 model. Meanwhile, for the OAB Exams (Brazilian Bar Exam), performance appears entirely uncorrelated with the number of tokens processed, despite over 4 billion tokens from our dataset originating from the legal domain, regardless of model size. We initially hypothesized that model performance might only exceed random chance for benchmarks like ENEM, BLUEX, and OAB Exams once the models surpass a certain parameter threshold (e.g., 7 billion), which would explain the poor performance of smaller models. However, this does not account for why performance correlates significantly with training volume on similar benchmarks, such as ARC-Challenge and HellaSwag, which follow a multiple-choice Q\&A format.

At the same time, for all sub-billion parameter models, we find instances where \textit{"training hinders benchmark performance"}, i.e., inverse scaling. This is especially true for our 160 million parameter model, where, for several benchmarks, its performance worsens as the model advances its training. Also, for evaluations like HateBR and ASSIN2 STS, we again see this phenomenon afflicting our 2b4 model, where training causes the models to become worse than a random guesser. At the same time, performance on benchmarks like the Portuguese native FAQUAD NLI seems completely uncorrelated with token ingestion.

These results prompt us to question the validity of these evaluations and help explain other results presented in the literature\cite{lopes2024gl}. Regardless of the number of tokens in which models were trained, language modeling capabilities did not translate to performance in numerous evaluations used by the community. Hence, we hypothesize that results showing good performance on such benchmarks (i.e., above what a random guesser would achieve) might indicate not language modeling pretraining but overfitting to the style of evaluation these benchmarks bring (e.g., multiple choice Q\&A of OAB exams or ENEM tests), or simple luck\footnote{On benchmarks like ASSIN2 RTE, performance fluctuates drastically from checkpoint to checkpoint.}. However, given that many studies do not share the foundations of their work, like pretraining/fine-tuning datasets, it becomes hard to explain, on an empirical level, the performance documented by such developments \cite{pires2023sabi, almeida2024sabi, santos2024advancing}.

\begin{figure*}
    \centering
    \begin{subfigure}[b]{0.475\textwidth}
        \centering
        \includegraphics[width=\textwidth]{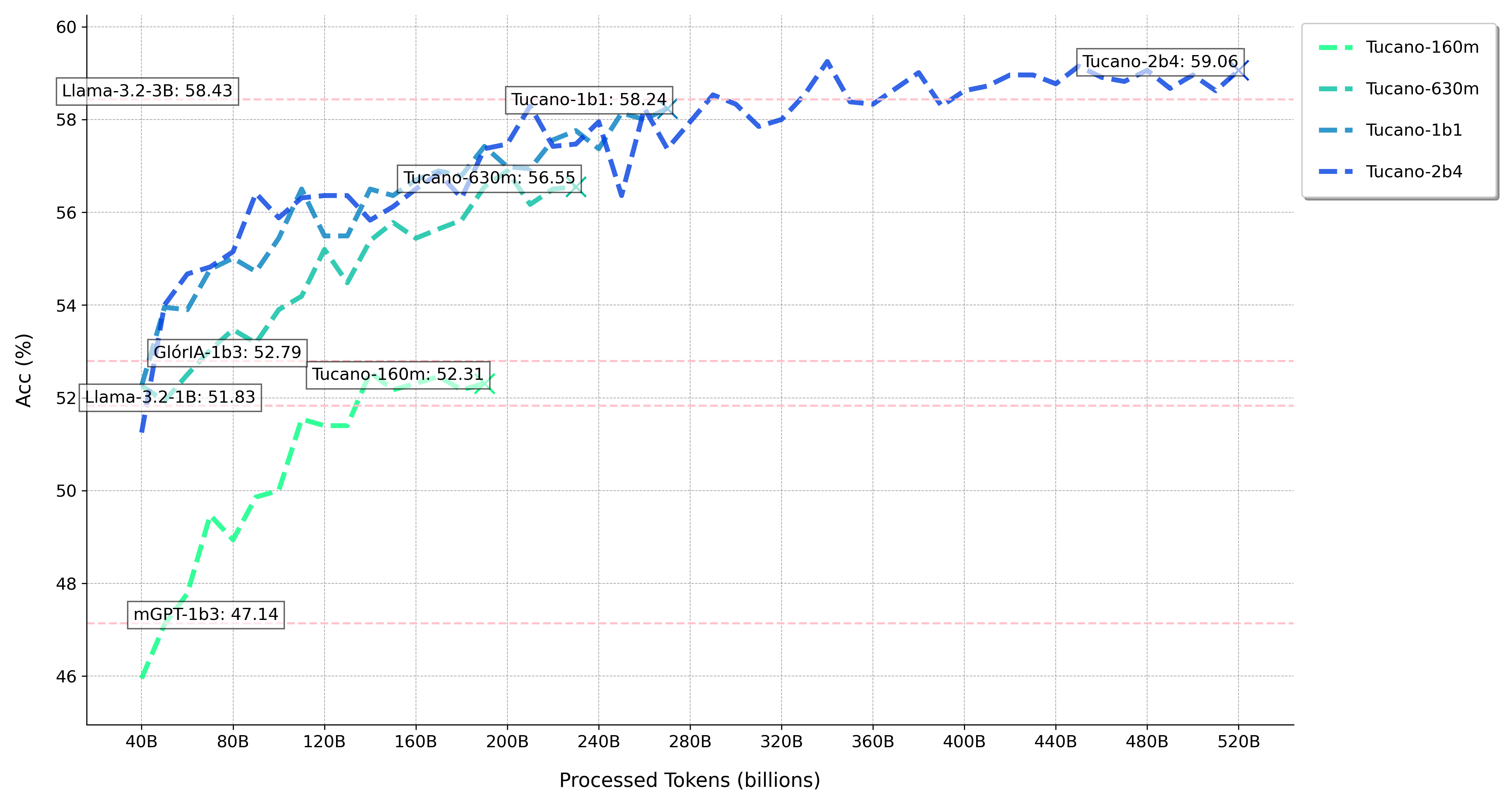}
        \caption[CALAME-PT]%
        {{\small CALAME-PT}}    
        \label{fig:calame}
    \end{subfigure}
    \hfill
    \begin{subfigure}[b]{0.475\textwidth}  
        \centering 
        \includegraphics[width=\textwidth]{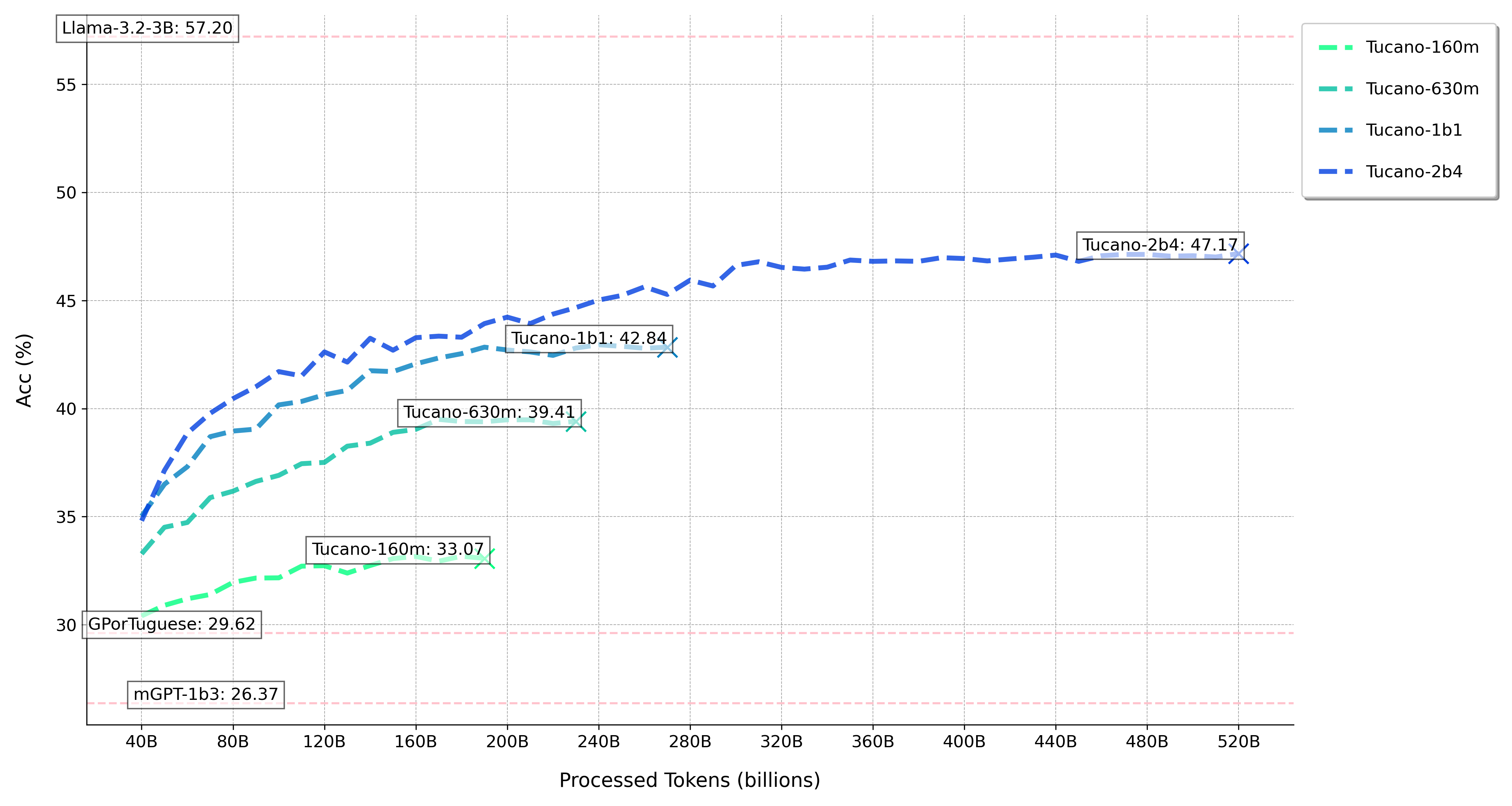}
        \caption[HellaSwag]%
        {{\small HellaSwag}}    
        \label{fig:hellaswag}
    \end{subfigure}
    \vskip\baselineskip
    \begin{subfigure}[b]{0.475\textwidth}   
        \centering 
        \includegraphics[width=\textwidth]{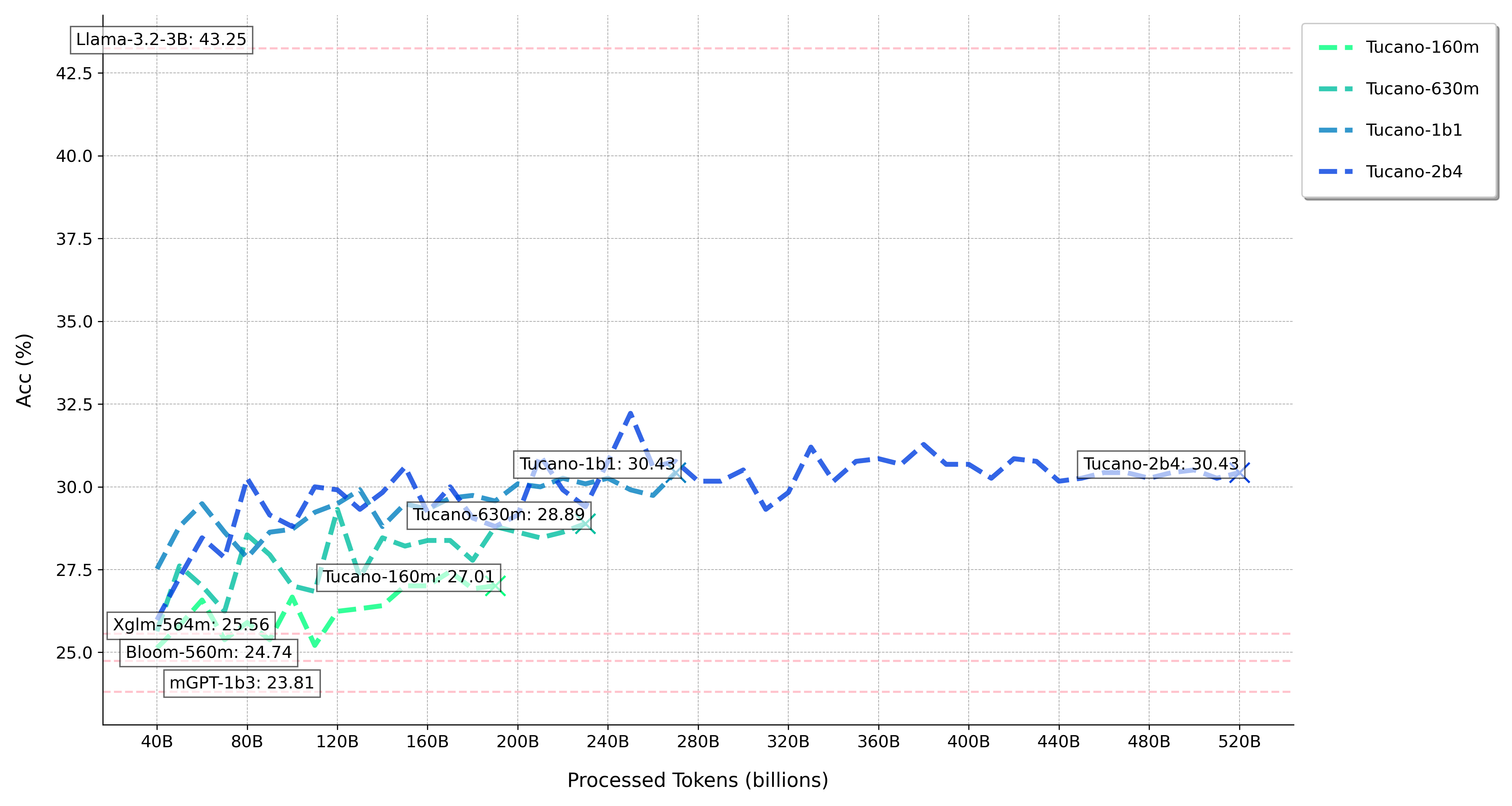}
        \caption[ARC-Challenge]%
        {{\small ARC-Challenge}}    
        \label{fig:arc}
    \end{subfigure}
    \hfill
    \begin{subfigure}[b]{0.475\textwidth}   
        \centering 
        \includegraphics[width=\textwidth]{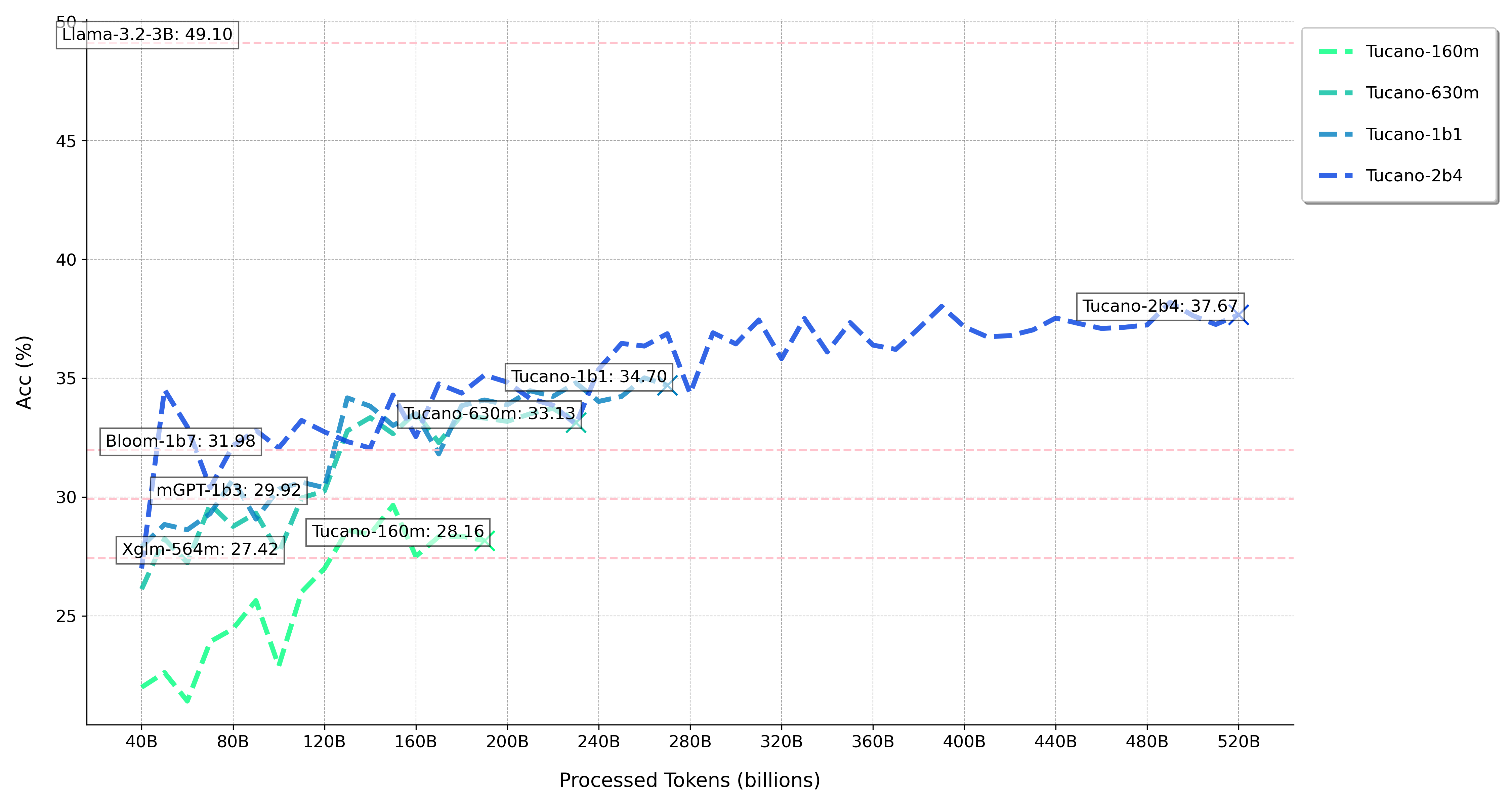}
        \caption[LAMBADA]%
        {{\small LAMBADA}}    
        \label{fig:lambada}
    \end{subfigure}
    \caption[ CALAME-PT, HellaSwag, ARC-Challenge, LAMBADA ]
    {\small The images above show instances where benchmark performance increases as models are trained on more tokens. The same analysis for all benchmarks used in this study can be found in our \href{https://github.com/Nkluge-correa/Tucano/tree/main/logs/README.md}{GitHub} repository.} 
    \label{fig:benchmarks}
\end{figure*}

Despite these findings, we observed several evaluations where the extent of language modeling pretraining shows an above-average (> 60\%) positive correlation with benchmark performance across the entire series (Fig. \ref{fig:benchmarks}). Benchmarks such as CALAME-PT (\ref{fig:calame}), LAMBADA(\ref{fig:lambada}), HellaSwag(\ref{fig:hellaswag}), and the ARC-Challenge(\ref{fig:arc}) consistently showed improvement as causal language modeling pretraining scales. These benchmarks, therefore, seem to serve as the most reliable indicators of model performance and capabilities when training native Portuguese LLMs with plain common crawl data. These insights could assist other practitioners in selecting benchmarks for evaluating smaller LLMs and determining which benchmarks might be better suited for models specifically trained or fine-tuned for particular domains (e.g., OAB test exams). 

Focusing only on the benchmarks that showed a significant correlation between language modeling pretraining and performance, we obtained the results in Table \ref{tab:benchmark-portuguese-language-models}. According to our evaluation protocol, our largest models outperformed several multilingual and natively pre-trained LLMs across nearly all benchmarks, including the recently released Llama-3.2-1b, trained on a far larger dataset than GigaVerbo. Our models also outperformed larger multilingual models, such as Bloom-1b7, in benchmarks like CALAME-PT and LAMBADA. Considering all benchmarks in our evaluation suite, our series outperforms all models listed in Table \ref{tab:benchmark-portuguese-language-models}, except the models coming from the Llama-3.2 series \footnote{Additionally, we noticed that Llama-based models like Sabiá-7b, Gervásio-7b, and Llama-2-7b significantly outperform other models on benchmarks where we observed a low correlation between language modeling pretraining and evaluation performance (e.g., ENEM, BLUEX, OAB Exams, FAQUAD NLI). Since we do not have access to the training data for these models (particularly the pretraining corpus of the Llama series), we suspect this performance discrepancy may be due to overfitting on specific evaluation styles, which, we speculate, could require particular types of data or domain knowledge to achieve the documented results.}.

\begin{table*}[h]
  \centering
  \small
  \begin{tabular}{lccccc}
    \toprule
     & \textbf{Average} & \textbf{Calame-PT} & \textbf{Lambada-PT} & \textbf{ARC-PT} & \textbf{HellaSwag-PT} \\
    \midrule
    Llama-3.2-3B       & 52.00  & 58.43     & 49.1      & 43.25  & 57.2   \\
    \textbf{Tucano-2b4} & 43.58  & 59.06     & 37.67     & 30.43  & 47.17  \\
    Llama-3.2-1B       & 42.95  & 51.83     & 41.02     & 33.5   & 45.44  \\
    \textbf{Tucano-1b1} & 41.55  & 58.24     & 34.7      & 30.43  & 42.84  \\
    Gemma-2b           & 40.38  & 51.16     & 39.88     & 37.95  & 32.53  \\
    Bloom-1b7          & 40.37  & 55.64     & 31.98     & 30.34  & 43.52  \\
    \textbf{Tucano-630m} & 39.5   & 56.55     & 33.13     & 28.89  & 39.41  \\
    Gemma-2-2b         & 39.21  & 56.7      & 47.1      & 24.19  & 28.85  \\
    Bloom-1b1          & 38.18  & 52.94     & 30.22     & 29.83  & 39.74  \\
    GlórIA-1b3         & 36.05  & 52.79     & 27.71     & 26.67  & 37.04  \\
    \textbf{Tucano-160m} & 35.14  & 52.31     & 28.16     & 27.01  & 33.07  \\
    XGLM-564m          & 34.55  & 50.58     & 27.42     & 25.56  & 34.64  \\
    Bloom-560m         & 34.32  & 49.95     & 25.44     & 24.74  & 37.15  \\
    TTL-460m           & 33.78  & 49.42     & 23.29     & 29.4   & 33.00  \\
    mGPT-1b3           & 31.81  & 47.14     & 29.92     & 23.81  & 26.37  \\
    TTL-160m           & 30.78  & 46.72     & 20.98     & 26.15  & 29.29  \\
    Lola-v1            & 30.19  & 26.4      & 18.32     & 30.42  & 45.61  \\
    GPorTuguese        & 28.92  & 40.61     & 22.98     & 22.48  & 29.62  \\
    \bottomrule
  \end{tabular}
  \vspace{0.25cm}
  \caption{Evaluation benchmark scores for our models compared with models of similar size. For this table, we use only the benchmarks that demonstrated a positive correlation (> 60\%) between benchmark performance and token ingestion across the entire series. All evaluations for all benchmarks that form our custom harness are available on our \href{https://github.com/Nkluge-correa/Tucano/blob/main/evaluations/README.md}{GitHub} repository.}
  \label{tab:benchmark-portuguese-language-models}
\end{table*}

It is also noteworthy that the fine-tuning/alignment process has the potential to degrade the performance of the foundational model on specific benchmarks. For instance, while our alignment process improved the controllability of our models for users, it reduced their performance in particular benchmarks\footnote{The same applies to Gervásio-7b, which performs considerably worse than its raw base (Llama-2-7b) across nearly all evaluations.}. However, when looking at our Instruct models via our custom AlpacaEval benchmark, a more appropriate benchmark for evaluating chat models, we see promising results (Table \ref{tab:length-controlled-win-rates}), where Tucano-Instruct models can outperform much larger models (e.g., Sabiá-7b and Gervásio-7b), and approximate models like the ones from the Llama-3.2 series.

\begin{table*}[h]
  \centering
  \small
  \begin{tabular}{lcccccc}
    \toprule
    & \textbf{Avg. Length} & \textbf{Wins} & \textbf{Base Wins} & \textbf{LC Win Rate (\%)} & \textbf{Std. Error} \\
    \midrule
    Llama-3.2-3B-Instruct        & 1609  & 257 & 548 & 21.06 & 0.075 \\
    \textbf{Tucano-2b4-Instruct} & 1843  & 151 & 654 & 13.00 & 0.071 \\
    \textbf{Tucano-1b1-Instruct} & 1667  & 124 & 681 & 8.80  & 0.083 \\
    Llama-3.2-1B-Instruct        & 1429  & 99  & 706 & 7.15  & 0.057 \\
    TeenyTinyLlama-460m-Chat     & 1333  & 28  & 777 & 2.84  & 0.059 \\
    Sabiá-7b                     & 5011  & 1   & 804 & 0.076 & 0.0043 \\
    Gervásio-7b                  & 5740  & 1   & 804 & 0.026 & 0.0016 \\
    \bottomrule
  \end{tabular}
  \vspace{0.25cm}
  \caption{Length-controlled win rates and standard errors for different models evaluated with the \href{https://huggingface.co/datasets/TucanoBR/alpaca-eval-pt}{Alpaca-Eval-PT}. All models were evaluated on 805 prompts. To replicate our evaluation, visit our \href{https://github.com/Nkluge-correa/Tucano/blob/main/evaluations/README.md}{GitHub} repository.}
  \label{tab:length-controlled-win-rates}
\end{table*}

\subsection{Qualitative Demonstration}
\label{section8-sub-2}

To complement our evaluations, Table \ref{tab:qualitative} presents a qualitative demonstration of Tucano's text generation capabilities by prompting our largest models on diverse topics related to Brazilian and Portuguese culture.

\begin{table}[h!]
    \centering
    \begin{tabular}{p{0.5\linewidth} | p{0.5\linewidth}}

        \toprule
        
        \textcolor{ForestGreen}{A floresta da Amazônia é conhecida por sua} \textit{biodiversidade e diversidade de espécies vegetais e animais. No entanto, a exploração madeireira ilegal e o desmatamento continuam a ameaçar esta importante área natural. A extração ilegal de madeira na Amazônia tem sido um problema significativo. Um estudo realizado pelo World Wildlife Fund (WWF) em 2019 descobriu que mais de 80\% das árvores derrubadas na região foram ilegalmente extraídas. O desmatamento, juntamente com as mudanças climáticas, está causando uma perda significativa de habitat para muitas espécies, incluindo pássaros, mamíferos e insetos. (...)} & \textcolor{ForestGreen}{O Carnaval do Rio de Janeiro é} \textit{um dos mais famosos e tradicionais do mundo. A festa acontece durante os dias que antecedem a Quarta-Feira de Cinzas, o primeiro dia da Quaresma. Durante cinco dias, os foliões saem às ruas em busca de diversão e adoram celebrar até o amanhecer nas festas de rua e nos desfiles das escolas de samba. A cidade do Rio de Janeiro, conhecida mundialmente por suas belezas naturais, também tem seu carnaval. As pessoas se reúnem para curtir as marchinhas e sambas-enredo das escolas de samba e os blocos de rua (...)} \\

        \midrule
        
        \textcolor{ForestGreen}{<instruction>Qual é a capital do Estado do Rio Grande do Sul?</instruction>}\textit{A capital do estado do Rio Grande do Sul, Brasil, é Porto Alegre. É um importante centro cultural, econômico e político no sul do Brasil. A cidade tem uma população de cerca de 1,4 milhão de habitantes e está localizada na região metropolitana de Porto Alegre, que inclui cidades vizinhas como Canoas, Gravataí e Viamão (...)} & \textcolor{ForestGreen}{"<instruction>Quem foi Fernando Pessoa?</instruction>"}\textit{Fernando Pessoa foi um poeta, escritor e tradutor português nascido em 13 de junho de 1888, em Lisboa. Ele é amplamente considerado como o maior poeta da língua portuguesa e uma figura cultural significativa do século XX. Aqui estão alguns detalhes sobre sua vida e obra: (...)} \\

        \midrule

        \textcolor{ForestGreen}{A culinária Portuguesa é famosa por pratos como o} \textit{bacalhau, a alheira, os pastéis de nata e as sardinhas. Além disso, os vinhos portugueses são conhecidos mundialmente pela sua qualidade e sabor único. Se você está planejando uma viagem para Portugal, não deixe de experimentar esses pratos deliciosos e desfrutar da cultura gastronômica única do país (...)} & \textcolor{ForestGreen}{Um fator comun entre países como o Brasil, Angola e Moçambique é que todos foram} \textit{colonizados por Portugal e, portanto, têm uma história comum de colonização. O Brasil foi colônia portuguesa até 1822, quando a independência do país foi proclamada (...)} \\

        \bottomrule
    \end{tabular}

    \vspace{0.25cm}
    \caption{These text samples (\textcolor{ForestGreen}{prompts} and \textit{generations}) were generated by Tucano-2b4 (and 2b4-Instruct) using beam search decoding with the following sampling configuration: top-p = 1.0, temperature = 0.3, top-k = 100.}
    \label{tab:qualitative}
\end{table}

According to our initial explorations, Tucano models demonstrate a firm grasp of culturally relevant subjects tied to the Lusophone world and can generate coherent and contextually appropriate text regarding many subjects. However, like all LLMs, our models have strong tendencies towards generating hallucinations, i.e., text that is grammatically correct but factually erroneous or incomprehensible, besides other limitations tied to the fact that a significant portion of our pretraining corpus contains machine-translated samples of English text.

\subsection{Energy Consumption and Carbon Emissions}
\label{section8-sub-3}

Following the example of past works \cite{strubell2019energy, garcia2019estimation, lottick2019energy, lacoste2019quantifying, desislavov2021compute, luccioni2022estimating}, and with the knowledge of the side effects deep learning practices can have when scaled at the level required by large foundation models \cite{van2021sustainable, patterson2022carbon, falk2023challenging, falk2024attribution}, we tracked our energy use during training, experiments, and evaluation. We measured the energy consumption and estimated carbon emissions for every checkpoint created during our training runs and experiments. All estimations were made using the 2023 estimations of the carbon intensity of Germany's energy grid (0.37 KgCO$_2$eq/KWh)\footnote{Available in \href{https://github.com/mlco2/codecarbon/blob/master/codecarbon/data/private_infra/global_energy_mix.json}{CodeCarbon's} source \cite{codecarbon}.}, which, according to Lottick et al. \cite{lottick2019energy} methodology, can be used to infer carbon emissions (CO$_2$eq) by multiplying the carbon intensity of the energy grid by the total energy consumption of a given experiment. Table \ref{tab:energy-consumption} summarizes the energy and carbon footprint related to our work.

\begin{table}[htbp]
  \centering
  \begin{tabular}{c|cccc}
    \toprule
    \textbf{Model} & \textbf{Duration (hours)} & \textbf{Training (kWh)} & \textbf{Exp. (kWh)} &\textbf{Emissions (KgCO2eq)} \\
    \midrule
    160m & 44 & 235 & 200 & 160 \\[2mm]
    630m & 170 & 920 & 125 & 387 \\[2mm]
    1b1 & 194 & 2,600 & 335 & 1,085 \\[2mm]
    2b4 & 860 & 11,860 & 400 & 4,536 \\[2mm]
    \midrule
    \textbf{Total} & \textbf{1,268} & \textbf{15,615} & \textbf{1,060} & \textbf{6,168 KgCO2eq} \\
    \bottomrule
  \end{tabular}
 
  \vspace{0.25cm}
  \caption{The table above shows, for each model, the duration of its training run, the energy consumption related to that run, the energy consumption regarding experimentation and evaluations, and the total estimated carbon emissions regarding the development of that model size. The training of the instruct versions is also accounted for in each respective model. To minimize energy consumption, we performed almost all of our experiments using the smaller version of our models. According to our logs, we utilized around 5,900 GPU hours across training, translating to an estimated cost of approximately 5,990 USD, assuming a rate of 1.1 USD per hour per A100 GPU. From the total of 16,675 kWh used, a significant portion ($\approx$ 6\%) was used to run experiments and evaluations, totaling 6.1 tCO$_2$eq in emissions.}
  \label{tab:energy-consumption}
\end{table}

Deep learning research is fundamentally driven by experimentation and heuristic approaches. Although many studies attempt to document training procedures \cite{zhang2022opt, biderman2023pythiasuiteanalyzinglarge, dey2023cerebras, zhang2022opt}, offering valuable guidelines for configuring models and their training environments, these published (or documented) procedures rarely provide universal solutions. Hence, the heuristic challenges and the current deficiencies in training documentation force researchers to expend resources and energy that could have been avoided when developing new models. Meanwhile, several factors shape the carbon footprint of deep learning, including the unique characteristics of each experiment and the infrastructure supporting it. In our experience, we frequently needed to fine-tune hyperparameters, adjust preprocessing strategies, and conduct exploratory experiments to achieve good results. However, this reliance on experimentation has significant environmental implications. To address this issue, we performed most experiments using our smaller models, as experimenting with the larger models (e.g., 2b4) would have led to a much higher increase in CO$_2$ emissions, which we aimed to avoid. In short, LLM development is computationally demanding, with a substantial portion of energy consumption occurring outside the training runs.

\section{Future Works}
\label{section9}

The Tucano series significantly contributes to the Portuguese NLP community in several ways. First, we ensure that the entire series is open-source and highly reproducible. Additionally, the language models we present are trained on the largest documented dataset of native Portuguese text. To the best of our knowledge, the scale of monolingual Portuguese pretraining in this study is unprecedented in the literature. All models, along with intermediary checkpoints, datasets, code implementations, and logs, are freely accessible through the repositories associated with this study. Table \ref{tab:open-source-comparisons} summarizes the availability of the artifacts mentioned earlier in the context of the Portuguese LLMs reviewed in Section \ref{section2}, with a comparison to our own work.

Nevertheless, numerous milestones remain to be achieved before Portuguese can be considered a high-resource language. Some of the prospects for future studies are:

\begin{enumerate}
    \item \textbf{Expanding GigaVerbo by creating larger concatenation of Portuguese datasets}. Future studies should seek to enrich our pretraining corpus with more high-quality tokens, like academic papers, books, and other forms of high-quality text. Ambitiously, we should aim to reach the trillion-token range. At the same time, it would be interesting to conduct ablation studies on GigaVerbo to determine the impact of different dataset components and identify which subsets contribute most effectively to model performance.

    \item \textbf{Further enhancing GigaVerbo by incorporating synthetically generated data}. While this approach was not explored in our current study, synthetic data augmentation has been proven in other works to bolster model performance in many specific domains (e.g., coding and storytelling) \cite{gunasekar2023textbooks}. In the future, augmenting GigaVerbo with this type of data could improve its representative power in domains where, in its current state, it is found to be lacking.

    \item{Explore the downstream applicability of the Tucano series}: Future studies can use the models from the Tucano series as foundations for future developments, like multimodal Portuguese LLaVas \cite{liu2023visualinstructiontuning}, Portuguese embedding models \cite{behnamghader2024llm2vec}, or more capable filters and guardrails.
    
    \item \textbf{Increasing model scale to larger architectures, such as 3B, 7B, and 13B parameters}. Scaling up to larger model sizes would enable us to understand better how benchmark performance changes with model size and to determine whether certain benchmarks correlate more strongly with language modeling pretraining only when models exceed a certain size threshold.
    
    \item \textbf{Developing new and more comprehensive benchmarks for Portuguese}. Our results indicate that Portuguese evaluation benchmarks for generative language models require improvement. Future research to advance Portuguese NLP should focus on either developing more effective benchmarks or refining existing ones to better capture the impact of pretraining and provide a more precise correlation between pretraining depth and performance across various language tasks.
\end{enumerate}

\begin{table*}[h]
  \centering
  \scalebox{1.0}{
  \begin{tabular}{p{0.4\linewidth} | cccccc}
    \toprule
      & \textbf{Model} & \textbf{Data} & \textbf{Code} & \textbf{Logs} & \textbf{\#models} & \textbf{\#ckpts}\\
       
    \midrule

    \textbf{\textcolor{ForestGreen}{Tu}\textcolor{Dandelion}{ca}\textcolor{RoyalBlue}{no}} & \textcolor{Green}{\checkmark} & \textcolor{Green}{\checkmark} & \textcolor{Green}{\checkmark} & \textcolor{Green}{\checkmark} & 6 & 111 \\
    
    TeenyTinyLlama & \textcolor{Green}{\checkmark} & \textcolor{Green}{\checkmark} & \textcolor{Green}{\checkmark} & \textcolor{Green}{\checkmark} & 3 & 70 \\

    GPorTuguese & \textcolor{Green}{\checkmark} & \textcolor{Green}{\checkmark} & \textcolor{Green}{\checkmark} & \textcolor{Green}{\checkmark} & 1 & 1 \\

    PTT5 & \textcolor{Green}{\checkmark} & \textcolor{Green}{\checkmark} & \textcolor{Green}{\checkmark} & \textcolor{Red}{\xmark} & 6 & 1 \\

    RoBERTaLexPT & \textcolor{Green}{\checkmark} & \textcolor{Green}{\checkmark} & \textcolor{Red}{\xmark} & \textcolor{Red}{\xmark} & 2 & 3 \\

    Albertina & \textcolor{Green}{\checkmark} & \textcolor{Green}{\checkmark} & \textcolor{Red}{\xmark} & \textcolor{Red}{\xmark} & 8 & 1 \\

    BERTimbau & \textcolor{Green}{\checkmark} & \textcolor{Green}{\checkmark} & \textcolor{Red}{\xmark} & \textcolor{Red}{\xmark} & 2 & 1 \\

    DeBERTinha & \textcolor{Green}{\checkmark} & \textcolor{Green}{\checkmark} & \textcolor{Red}{\xmark} & \textcolor{Red}{\xmark} & 1 & 1 \\

    Gervásio & \textcolor{Green}{\checkmark} & \textcolor{Green}{\checkmark} & \textcolor{Red}{\xmark} & \textcolor{Red}{\xmark} & 2 & 1 \\

    PTT5-v2 & \textcolor{Green}{\checkmark} & \textcolor{Red}{\xmark} & \textcolor{Red}{\xmark} & \textcolor{Red}{\xmark} & 4 & 1 \\

    BERTabaporu & \textcolor{Green}{\checkmark} & \textcolor{Red}{\xmark} & \textcolor{Red}{\xmark} & \textcolor{Red}{\xmark} & 2 & 1 \\
    
    Glória & \textcolor{Green}{\checkmark} & \textcolor{Red}{\xmark} & \textcolor{Red}{\xmark} & \textcolor{Red}{\xmark} & 1 & 1 \\

    Sabiá & \textcolor{Green}{\checkmark} & \textcolor{Red}{\xmark} & \textcolor{Red}{\xmark} & \textcolor{Red}{\xmark} & 1 & 1 \\

    Sabiá-2 & \textcolor{Red}{\xmark} & \textcolor{Red}{\xmark} & \textcolor{Red}{\xmark} & \textcolor{Red}{\xmark} & 2 & None \\

    Sabiá-3 & \textcolor{Red}{\xmark} & \textcolor{Red}{\xmark} & \textcolor{Red}{\xmark} & \textcolor{Red}{\xmark} & 1 & None \\

    \bottomrule
  \end{tabular}
  }
  \vspace{0.25cm}
  \caption{The above table compares Portuguese language models regarding the open-source availability of models, datasets, code, logs, the total number of models (\#models), and the number of checkpoints (\#ckpts). In terms of open (and reproducible) development, many aspects of past studies are indeed closed. Saved for rare exceptions \cite{pierre2020gpt2smallportuguese, carmo2020ptt5, correa2024teenytinyllama}, many studies only make available "end-products" devoid of logs, datasets, or code implementations, making the reproduction of LLM development a task that requires constant rediscovering. Given the level of computing needed to practice deep learning at such scales, a lack of reusable code and materials can seriously slow down the Portuguese NLP community's progress while hindering its sustainability.}
  \label{tab:open-source-comparisons}
\end{table*}

\section{Conclusion}
\label{section10}

In this study, we introduced the \textbf{\textcolor{ForestGreen}{Tu}\textcolor{Dandelion}{ca}\textcolor{RoyalBlue}{no}} series, a collection of open-source large language models designed to advance natural language processing for Portuguese. Our work covered the entire development pipeline, from dataset creation and filtration to hyperparameter tuning and evaluation, emphasizing openness and reproducibility. These efforts contribute capable models, large datasets, and tools to the Portuguese NLP community to set a standard for transparent and replicable research practices. Moreover, our critical assessment of the field highlighted ongoing challenges, particularly around evaluation methodologies and result interpretability, which will only be solved if the community shifts toward a more rigorous and reproducible developmental framework. Finally, we hope our contributions will help spur this shift, providing essential resources to guide future studies. Ultimately, we hope the work initiated here will be extended to other low-resource languages, fostering a more equitable and sustainable NLP ecosystem globally.

\section*{Acknowledgments}

The authors gratefully acknowledge the granted access to the \href{https://www.hpc.uni-bonn.de/en/systems/marvin}{Marvin cluster} hosted by the \href{https://www.uni-bonn.de/en}{University of Bonn} along with the support provided by its High Performance Computing \& Analytics Lab. Authors would also like to acknowledge their own personal funding agencies. Nicholas Kluge Corrêa is funded by the Ministerium für Wirtschaft, Industrie, Klimaschutz und Energie des Landes Nordrhein-Westfalen (Ministry for Economic Affairs, Industry, Climate Action and Energy of the State of North Rhine- Westphalia), as part of the KI.NRW-flagship project "\href{https://www.zertifizierte-ki.de/}{Zertifizierte KI}" (Certified AI). Aniket Sen is funded by the Deutsche Forschungsgemeinschaft (DFG, German Research Foundation) as part of the CRC 1639 \href{https://numeriqs.hiskp.uni-bonn.de/}{NuMeriQS} – project no. 511713970.

\section*{Author's Information}

The corresponding author is \textbf{Nicholas Kluge Corrêa}. He is a postdoctoral researcher at the Center for Science and Thought at the University of Bonn (Bonn, NRW, Germany). His contact email is \href{mailto:kluge@uni-bonn.de}{kluge@uni-bonn.de}.

\textbf{Aniket Sen} is a postdoctoral researcher at the High Performance Computing and Analytics Lab and the Helmholtz-Institut für Strahlen- und Kernphysik at the University of Bonn. His contact email is \href{mailto:sen@hiskp.uni-bonn.de}{sen@hiskp.uni-bonn.de}.

\textbf{Sophia Falk} is a PhD researcher at the Bonn Sustainable AI Lab, Institute for Science and Ethics, University of Bonn. Her contact email is \href{mailto:falk@iwe.uni-bonn.de}{falk@iwe.uni-bonn.de}.

\textbf{Shiza Fatimah} is a master's student at the Institute for Computer Science at the University of Bonn. Her contact email is \href{mailto:s39sfati@uni-bonn.de}{s39sfati@uni-bonn.de}.

\bibliographystyle{plain}
\bibliography{references}

\begin{thebibliography}{100}

\bibitem{abadji2022towards}
Julien Abadji, Pedro~Ortiz Suarez, Laurent Romary, and Beno{\^\i}t Sagot.
\newblock Towards a cleaner document-oriented multilingual crawled corpus.
\newblock {\em arXiv preprint arXiv:2201.06642}, 2022.

\bibitem{adams2017cross}
Oliver Adams, Adam Makarucha, Graham Neubig, Steven Bird, and Trevor Cohn.
\newblock Cross-lingual word embeddings for low-resource language modeling.
\newblock In {\em Proceedings of the 15th Conference of the European Chapter of the Association for Computational Linguistics: Volume 1, Long Papers}, pages 937--947, 2017.

\bibitem{ainslie2023gqa}
Joshua Ainslie, James Lee-Thorp, Michiel de~Jong, Yury Zemlyanskiy, Federico Lebr{\'o}n, and Sumit Sanghai.
\newblock Gqa: Training generalized multi-query transformer models from multi-head checkpoints.
\newblock {\em arXiv preprint arXiv:2305.13245}, 2023.

\bibitem{almazrouei2023falcon}
Ebtesam Almazrouei, Hamza Alobeidli, Abdulaziz Alshamsi, Alessandro Cappelli, Ruxandra Cojocaru, M{\'e}rouane Debbah, {\'E}tienne Goffinet, Daniel Hesslow, Julien Launay, Quentin Malartic, et~al.
\newblock The falcon series of open language models.
\newblock {\em arXiv preprint arXiv:2311.16867}, 2023.

\bibitem{almeida2024sabi}
Thales~Sales Almeida, Hugo Abonizio, Rodrigo Nogueira, and Ramon Pires.
\newblock Sabi$\backslash$'a-2: A new generation of portuguese large language models.
\newblock {\em arXiv preprint arXiv:2403.09887}, 2024.

\bibitem{almeida2023bluex}
Thales~Sales Almeida, Thiago Laitz, Giovana~K. Bonás, and Rodrigo Nogueira.
\newblock Bluex: A benchmark based on brazilian leading universities entrance exams, 2023.

\bibitem{anand2023gpt4all}
Yuvanesh Anand, Zach Nussbaum, Brandon Duderstadt, Benjamin Schmidt, and Andriy Mulyar.
\newblock Gpt4all: Training an assistant-style chatbot with large scale data distillation from gpt-3.5-turbo.
\newblock {\em GitHub (2023)}, 2023.

\bibitem{armengol2021multilingual}
Jordi Armengol-Estap{\'e}, Casimiro~Pio Carrino, Carlos Rodriguez-Penagos, Ona de~Gibert Bonet, Carme Armentano-Oller, Aitor Gonzalez-Agirre, Maite Melero, and Marta Villegas.
\newblock Are multilingual models the best choice for moderately under-resourced languages? a comprehensive assessment for catalan.
\newblock {\em arXiv preprint arXiv:2107.07903}, 2021.

\bibitem{arnold2005europe}
David Arnold.
\newblock Europe, technology, and colonialism in the 20th century.
\newblock {\em History and Technology}, 21(1):85--106, 2005.

\bibitem{aryabumi2024aya}
Viraat Aryabumi, John Dang, Dwarak Talupuru, Saurabh Dash, David Cairuz, Hangyu Lin, Bharat Venkitesh, Madeline Smith, Kelly Marchisio, Sebastian Ruder, et~al.
\newblock Aya 23: Open weight releases to further multilingual progress.
\newblock {\em arXiv preprint arXiv:2405.15032}, 2024.

\bibitem{bahdanau2014neural}
Dzmitry Bahdanau, Kyunghyun Cho, and Yoshua Bengio.
\newblock Neural machine translation by jointly learning to align and translate.
\newblock {\em arXiv preprint arXiv:1409.0473}, 2014.

\bibitem{behnamghader2024llm2vec}
Parishad BehnamGhader, Vaibhav Adlakha, Marius Mosbach, Dzmitry Bahdanau, Nicolas Chapados, and Siva Reddy.
\newblock Llm2vec: Large language models are secretly powerful text encoders.
\newblock {\em arXiv preprint arXiv:2404.05961}, 2024.

\bibitem{bengio2000neural}
Yoshua Bengio, R{\'e}jean Ducharme, and Pascal Vincent.
\newblock A neural probabilistic language model.
\newblock {\em Advances in neural information processing systems}, 13, 2000.

\bibitem{biderman2023pythiasuiteanalyzinglarge}
Stella Biderman, Hailey Schoelkopf, Quentin Anthony, Herbie Bradley, Kyle O'Brien, Eric Hallahan, Mohammad~Aflah Khan, Shivanshu Purohit, USVSN~Sai Prashanth, Edward Raff, Aviya Skowron, Lintang Sutawika, and Oskar van~der Wal.
\newblock Pythia: A suite for analyzing large language models across training and scaling, 2023.

\bibitem{biderman2023pythia}
Stella Biderman, Hailey Schoelkopf, Quentin~Gregory Anthony, Herbie Bradley, Kyle O’Brien, Eric Hallahan, Mohammad~Aflah Khan, Shivanshu Purohit, USVSN~Sai Prashanth, Edward Raff, et~al.
\newblock Pythia: A suite for analyzing large language models across training and scaling.
\newblock In {\em International Conference on Machine Learning}, pages 2397--2430. PMLR, 2023.

\bibitem{black2022gpt}
Sid Black, Stella Biderman, Eric Hallahan, Quentin Anthony, Leo Gao, Laurence Golding, Horace He, Connor Leahy, Kyle McDonell, Jason Phang, et~al.
\newblock Gpt-neox-20b: An open-source autoregressive language model.
\newblock {\em arXiv preprint arXiv:2204.06745}, 2022.

\bibitem{bommasani2021opportunities}
Rishi Bommasani, Drew~A Hudson, Ehsan Adeli, Russ Altman, Simran Arora, Sydney von Arx, Michael~S Bernstein, Jeannette Bohg, Antoine Bosselut, Emma Brunskill, et~al.
\newblock On the opportunities and risks of foundation models.
\newblock {\em arXiv preprint arXiv:2108.07258}, 2021.

\bibitem{bonifacio2020study}
Luiz~Henrique Bonifacio, Paulo~Arantes Vilela, Gustavo~Rocha Lobato, and Eraldo~Rezende Fernandes.
\newblock A study on the impact of intradomain finetuning of deep language models for legal named entity recognition in portuguese.
\newblock In {\em Intelligent Systems: 9th Brazilian Conference, BRACIS 2020, Rio Grande, Brazil, October 20--23, 2020, Proceedings, Part I 9}, pages 648--662. Springer, 2020.

\bibitem{brown2020language}
Tom Brown, Benjamin Mann, Nick Ryder, Melanie Subbiah, Jared~D Kaplan, Prafulla Dhariwal, Arvind Neelakantan, Pranav Shyam, Girish Sastry, Amanda Askell, et~al.
\newblock Language models are few-shot learners.
\newblock {\em Advances in neural information processing systems}, 33:1877--1901, 2020.

\bibitem{brum2017building}
Henrico~Bertini Brum and Maria das Gra{\c{c}}as~Volpe Nunes.
\newblock Building a sentiment corpus of tweets in brazilian portuguese.
\newblock {\em arXiv preprint arXiv:1712.08917}, 2017.

\bibitem{henrique2023caramelo}
{Bruno Henrique}.
\newblock Caramelo 7b.
\newblock \url{https://huggingface.co/Bruno/Caramelo_7B}, 2023.

\bibitem{henrique2023harpia}
{Bruno Henrique}.
\newblock Harpia-7b-guanacolora.
\newblock \url{https://huggingface.co/Bruno/Harpia-7b-guanacoLora}, 2023.

\bibitem{campiotti2023debertinha}
Israel Campiotti, Matheus Rodrigues, Yuri Albuquerque, Rafael Azevedo, and Alyson Andrade.
\newblock Debertinha: A multistep approach to adapt debertav3 xsmall for brazilian portuguese natural language processing task, 2023.

\bibitem{moro2024dataset2}
{Carlo Moro}.
\newblock Gpt4-500k-augmented-ptbr-clean.
\newblock \url{https://huggingface.co/datasets/cnmoro/GPT4-500k-Augmented-PTBR-Clean}, 2024.

\bibitem{moro2024dataset}
{Carlo Moro}.
\newblock Instruct-ptbr-enus-11m.
\newblock \url{https://huggingface.co/datasets/cnmoro/Instruct-PTBR-ENUS-11M}, 2024.

\bibitem{carmo2020ptt5}
Diedre Carmo, Marcos Piau, Israel Campiotti, Rodrigo Nogueira, and Roberto Lotufo.
\newblock Ptt5: Pretraining and validating the t5 model on brazilian portuguese data.
\newblock {\em arXiv preprint arXiv:2008.09144}, 2020.

\bibitem{chen2023chinesewebtext}
Jianghao Chen, Pu~Jian, Tengxiao Xi, Dongyi Yi, Qianlong Du, Chenglin Ding, Guibo Zhu, Chengqing Zong, Jinqiao Wang, and Jiajun Zhang.
\newblock Chinesewebtext: Large-scale high-quality chinese web text extracted with effective evaluation model, 2023.

\bibitem{chen2016xgboost}
Tianqi Chen and Carlos Guestrin.
\newblock Xgboost: A scalable tree boosting system.
\newblock In {\em Proceedings of the 22nd acm sigkdd international conference on knowledge discovery and data mining}, pages 785--794, 2016.

\bibitem{chen2016training}
Tianqi Chen, Bing Xu, Chiyuan Zhang, and Carlos Guestrin.
\newblock Training deep nets with sublinear memory cost.
\newblock {\em arXiv preprint arXiv:1604.06174}, 2016.

\bibitem{chowdhery2022palm}
Aakanksha Chowdhery, Sharan Narang, Jacob Devlin, Maarten Bosma, Gaurav Mishra, Adam Roberts, Paul Barham, Hyung~Won Chung, Charles Sutton, Sebastian Gehrmann, et~al.
\newblock Palm: Scaling language modeling with pathways. arxiv 2022.
\newblock {\em arXiv preprint arXiv:2204.02311}, 10, 2022.

\bibitem{clark2018think}
Peter Clark, Isaac Cowhey, Oren Etzioni, Tushar Khot, Ashish Sabharwal, Carissa Schoenick, and Oyvind Tafjord.
\newblock Think you have solved question answering? try arc, the ai2 reasoning challenge.
\newblock {\em arXiv preprint arXiv:1803.05457}, 2018.

\bibitem{codecarbon}
CodeCarbon.
\newblock Codecarbon: Track emissions from compute and recommend ways to reduce their impact on the environment.
\newblock \url{https://github.com/mlco2/codecarbon}, 2019.

\bibitem{cohere2024gap}
{Cohere For AI team}.
\newblock Policy primer - the ai language gap.
\newblock \url{https://cohere.com/research/papers/policy-primer-the-ai-language-gap-2024-06-27}, 2024.

\bibitem{conneau-etal-2020-unsupervised}
Alexis Conneau, Kartikay Khandelwal, Naman Goyal, Vishrav Chaudhary, Guillaume Wenzek, Francisco Guzm{\'a}n, Edouard Grave, Myle Ott, Luke Zettlemoyer, and Veselin Stoyanov.
\newblock Unsupervised cross-lingual representation learning at scale.
\newblock In {\em Proceedings of the 58th Annual Meeting of the Association for Computational Linguistics}, pages 8440--8451, Online, July 2020. Association for Computational Linguistics.

\bibitem{conneau2020unsupervised}
Alexis Conneau, Kartikay Khandelwal, Naman Goyal, Vishrav Chaudhary, Guillaume Wenzek, Francisco Guzmán, Edouard Grave, Myle Ott, Luke Zettlemoyer, and Veselin Stoyanov.
\newblock Unsupervised cross-lingual representation learning at scale, 2020.

\bibitem{DatabricksBlog2023DollyV2}
Mike Conover, Matt Hayes, Ankit Mathur, Jianwei Xie, Jun Wan, Sam Shah, Ali Ghodsi, Patrick Wendell, Matei Zaharia, and Reynold Xin.
\newblock Free dolly: Introducing the world's first truly open instruction-tuned llm, 2023.

\bibitem{correa2024dynamic}
Nicholas~Kluge Corr{\^e}a.
\newblock Dynamic normativity: Necessary and sufficient conditions for value alignment.
\newblock {\em arXiv preprint arXiv:2406.11039}, 2024.

\bibitem{correa2024teenytinyllama}
Nicholas~Kluge Corr{\^e}a, Sophia Falk, Shiza Fatimah, Aniket Sen, and Nythamar De~Oliveira.
\newblock Teenytinyllama: open-source tiny language models trained in brazilian portuguese.
\newblock {\em Machine Learning with Applications}, 16:100558, 2024.

\bibitem{costa2023bertabaporu}
Pablo~Botton Costa, Matheus~Camasmie Pavan, Wesley~Ramos Santos, Samuel~Caetano Silva, and Ivandr{\'e} Paraboni.
\newblock Bertabaporu: assessing a genre-specific language model for portuguese nlp.
\newblock In {\em Proceedings of the 14th International Conference on Recent Advances in Natural Language Processing}, pages 217--223, 2023.

\bibitem{cruz2019evaluating}
Jan Christian~Blaise Cruz and Charibeth Cheng.
\newblock Evaluating language model finetuning techniques for low-resource languages.
\newblock {\em arXiv preprint arXiv:1907.00409}, 2019.

\bibitem{cui2023efficient}
Yiming Cui, Ziqing Yang, and Xin Yao.
\newblock Efficient and effective text encoding for chinese llama and alpaca.
\newblock {\em arXiv preprint arXiv:2304.08177}, 2023.

\bibitem{dao2023flashattention2}
Tri Dao.
\newblock Flash{A}ttention-2: Faster attention with better parallelism and work partitioning.
\newblock In {\em International Conference on Learning Representations (ICLR)}, 2024.

\bibitem{dao2022flashattention}
Tri Dao, Daniel~Y. Fu, Stefano Ermon, Atri Rudra, and Christopher R{\'e}.
\newblock Flash{A}ttention: Fast and memory-efficient exact attention with {IO}-awareness.
\newblock In {\em Advances in Neural Information Processing Systems (NeurIPS)}, 2022.

\bibitem{de2024new}
Ona De~Gibert, Graeme Nail, Nikolay Arefyev, Marta Ba{\~n}{\'o}n, Jelmer Van Der~Linde, Shaoxiong Ji, Jaume Zaragoza-Bernabeu, Mikko Aulamo, Gema Ram{\'\i}rez-S{\'a}nchez, Andrey Kutuzov, et~al.
\newblock A new massive multilingual dataset for high-performance language technologies.
\newblock {\em arXiv preprint arXiv:2403.14009}, 2024.

\bibitem{delfino2017passing}
Pedro Delfino, Bruno Cuconato, Edward~Hermann Haeusler, and Alexandre Rademaker.
\newblock Passing the brazilian oab exam: data preparation and some experiments.
\newblock {\em arXiv preprint arXiv:1712.05128}, 2017.

\bibitem{deng2018deep}
Li~Deng and Yang Liu.
\newblock {\em Deep learning in natural language processing}.
\newblock Springer, 2018.

\bibitem{desislavov2021compute}
Radosvet Desislavov, Fernando Mart{\'\i}nez-Plumed, and Jos{\'e} Hern{\'a}ndez-Orallo.
\newblock Compute and energy consumption trends in deep learning inference.
\newblock {\em arXiv preprint arXiv:2109.05472}, 2021.

\bibitem{devlin2018bert}
Jacob Devlin, Ming-Wei Chang, Kenton Lee, and Kristina Toutanova.
\newblock Bert: Pre-training of deep bidirectional transformers for language understanding.
\newblock {\em arXiv preprint arXiv:1810.04805}, 2018.

\bibitem{dey2023cerebras}
Nolan Dey, Gurpreet Gosal, Hemant Khachane, William Marshall, Ribhu Pathria, Marvin Tom, Joel Hestness, et~al.
\newblock Cerebras-gpt: Open compute-optimal language models trained on the cerebras wafer-scale cluster.
\newblock {\em arXiv preprint arXiv:2304.03208}, 2023.

\bibitem{dubey2024llama}
Abhimanyu Dubey, Abhinav Jauhri, Abhinav Pandey, Abhishek Kadian, Ahmad Al-Dahle, Aiesha Letman, Akhil Mathur, Alan Schelten, Amy Yang, Angela Fan, et~al.
\newblock The llama 3 herd of models.
\newblock {\em arXiv preprint arXiv:2407.21783}, 2024.

\bibitem{dubois2024length}
Yann Dubois, Bal{\'a}zs Galambosi, Percy Liang, and Tatsunori~B Hashimoto.
\newblock Length-controlled alpacaeval: A simple way to debias automatic evaluators.
\newblock {\em arXiv preprint arXiv:2404.04475}, 2024.

\bibitem{dubois2024alpacafarm}
Yann Dubois, Chen~Xuechen Li, Rohan Taori, Tianyi Zhang, Ishaan Gulrajani, Jimmy Ba, Carlos Guestrin, Percy~S Liang, and Tatsunori~B Hashimoto.
\newblock Alpacafarm: A simulation framework for methods that learn from human feedback.
\newblock {\em Advances in Neural Information Processing Systems}, 36, 2024.

\bibitem{elfwing2018sigmoid}
Stefan Elfwing, Eiji Uchibe, and Kenji Doya.
\newblock Sigmoid-weighted linear units for neural network function approximation in reinforcement learning.
\newblock {\em Neural networks}, 107:3--11, 2018.

\bibitem{falk2023challenging}
Sophia Falk and Aimee van Wynsberghe.
\newblock Challenging ai for sustainability: what ought it mean?
\newblock {\em AI and Ethics}, pages 1--11, 2023.

\bibitem{falk2024attribution}
Sophia Falk, Aimee van Wynsberghe, and Lisa Biber-Freudenberger.
\newblock The attribution problem of a seemingly intangible industry.
\newblock {\em Environmental Challenges}, page 101003, 2024.

\bibitem{feng2020language}
Fangxiaoyu Feng, Yinfei Yang, Daniel Cer, Naveen Arivazhagan, and Wei Wang.
\newblock Language-agnostic bert sentence embedding.
\newblock {\em arXiv preprint arXiv:2007.01852}, 2020.

\bibitem{finardi2021berta}
Paulo Finardi, Jos{\'e}~Di{\'e} Viegas, Gustavo~T Ferreira, Alex~F Mansano, and Vinicius~F Carid{\'a}.
\newblock Berta$\backslash$'u: Ita$\backslash$'u bert for digital customer service.
\newblock {\em arXiv preprint arXiv:2101.12015}, 2021.

\bibitem{corpusCarolinaV1}
Marcelo Finger, Maria~Clara Paixão~de Sousa, Cristiane Namiuti, Vanessa Martins~do Monte, Aline~Silva Costa, Felipe~Ribas Serras, Mariana~Lourenço Sturzeneker, Raquel de~Paula Guets, Renata~Morais Mesquita, Guilherme Lamartine~de Mello, Maria Clara Ramos~Morales Crespo, Maria Lina de Souza~Jeannine Rocha, Patrícia Brasil, Mariana Marques~da Silva, and Mayara~Feliciano Palma.
\newblock Carolina: The open corpus for linguistics and artificial intelligence.
\newblock \url{ https://sites.usp.br/corpuscarolina/corpus}, 2022.
\newblock Version 1.1 (Ada).

\bibitem{fortuna2019hierarchically}
Paula Fortuna, Joao~Rocha da~Silva, Leo Wanner, S{\'e}rgio Nunes, et~al.
\newblock A hierarchically-labeled portuguese hate speech dataset.
\newblock In {\em Proceedings of the third workshop on abusive language online}, pages 94--104, 2019.

\bibitem{gandhe2014neural}
Ankur Gandhe, Florian Metze, and Ian Lane.
\newblock Neural network language models for low resource languages.
\newblock In {\em Fifteenth Annual Conference of the International Speech Communication Association}, 2014.

\bibitem{gao2020pile}
Leo Gao, Stella Biderman, Sid Black, Laurence Golding, Travis Hoppe, Charles Foster, Jason Phang, Horace He, Anish Thite, Noa Nabeshima, et~al.
\newblock The pile: An 800gb dataset of diverse text for language modeling.
\newblock {\em arXiv preprint arXiv:2101.00027}, 2020.

\bibitem{gao2021framework}
Leo Gao, Jonathan Tow, Stella Biderman, Sid Black, Anthony DiPofi, Charles Foster, Laurence Golding, Jeffrey Hsu, Kyle McDonell, Niklas Muennighoff, et~al.
\newblock A framework for few-shot language model evaluation.
\newblock {\em Version v0. 0.1. Sept}, 2021.

\bibitem{open-pt-llm-leaderboard}
Eduardo A.~S. Garcia.
\newblock Open portuguese llm leaderboard.
\newblock \url{https://huggingface.co/spaces/eduagarcia/open_pt_llm_leaderboard}, 2024.

\bibitem{garcia2024robertalexpt}
Eduardo~AS Garcia, Nadia~FF Silva, Felipe Siqueira, Hidelberg~O Albuquerque, Juliana~RS Gomes, Ellen Souza, and Eliomar~A Lima.
\newblock Robertalexpt: A legal roberta model pretrained with deduplication for portuguese.
\newblock In {\em Proceedings of the 16th International Conference on Computational Processing of Portuguese}, pages 374--383, 2024.

\bibitem{ultrachatBr}
Gabriel~Lino Garcia, Pedro~Henrique Paiola, João~Otávio Frediani, Luis~Henrique Morelli, João Vitor~Mariano Correia, Danilo~Samuel Jodas, Arnaldo~Candido Junior, Bruno~Elias Penteado, Ivan~Rizzo Guilherme, and João~Paulo Papa.
\newblock Ultrachatbr: Um dataset em português baseado no ultrachat, 2023.

\bibitem{bode2024}
Gabriel~Lino Garcia, Pedro~Henrique Paiola, Luis~Henrique Morelli, Giovani Candido, Arnaldo~Cândido Júnior, Danilo~Samuel Jodas, Luis C.~S. Afonso, Ivan~Rizzo Guilherme, Bruno~Elias Penteado, and João~Paulo Papa.
\newblock Introducing bode: A fine-tuned large language model for portuguese prompt-based task, 2024.

\bibitem{garcia2019estimation}
Eva Garc{\'\i}a-Mart{\'\i}n, Crefeda~Faviola Rodrigues, Graham Riley, and H{\aa}kan Grahn.
\newblock Estimation of energy consumption in machine learning.
\newblock {\em Journal of Parallel and Distributed Computing}, 134:75--88, 2019.

\bibitem{geiping2023cookbook}
Jonas Geiping, Quentin Garrido, Pierre Fernandez, Amir Bar, Hamed Pirsiavash, Yann LeCun, and Micah Goldblum.
\newblock A cookbook of self-supervised learning.
\newblock {\em arXiv preprint arXiv:2304.12210}, 2023.

\bibitem{openlm2023openllama}
Xinyang Geng and Hao Liu.
\newblock Openllama: An open reproduction of llama.
\newblock \url{https://github.com/openlm-research/open_llama}, May 2023.

\bibitem{goldman2024unpacking}
Omer Goldman, Avi Caciularu, Matan Eyal, Kris Cao, Idan Szpektor, and Reut Tsarfaty.
\newblock Unpacking tokenization: Evaluating text compression and its correlation with model performance.
\newblock {\em arXiv preprint arXiv:2403.06265}, 2024.

\bibitem{goodfellow2016deep}
Ian Goodfellow, Yoshua Bengio, and Aaron Courville.
\newblock {\em Deep learning}.
\newblock MIT press, 2016.

\bibitem{pierre2020gpt2smallportuguese}
Pierre Guillou.
\newblock Gportuguese-2 (portuguese gpt-2 small): a language model for portuguese text generation (and more nlp tasks...).
\newblock \url{https://huggingface.co/pierreguillou/gpt2-small-portuguese}, 2020.

\bibitem{gunasekar2023textbooks}
Suriya Gunasekar, Yi~Zhang, Jyoti Aneja, Caio C{\'e}sar~Teodoro Mendes, Allie Del~Giorno, Sivakanth Gopi, Mojan Javaheripi, Piero Kauffmann, Gustavo de~Rosa, Olli Saarikivi, et~al.
\newblock Textbooks are all you need.
\newblock {\em arXiv preprint arXiv:2306.11644}, 2023.

\bibitem{hasan-etal-2021-xl}
Tahmid Hasan, Abhik Bhattacharjee, Md.~Saiful Islam, Kazi Mubasshir, Yuan-Fang Li, Yong-Bin Kang, M.~Sohel Rahman, and Rifat Shahriyar.
\newblock {XL}-sum: Large-scale multilingual abstractive summarization for 44 languages.
\newblock In {\em Findings of the Association for Computational Linguistics: ACL-IJCNLP 2021}, pages 4693--4703, Online, August 2021. Association for Computational Linguistics.

\bibitem{hastie2009overview}
Trevor Hastie, Robert Tibshirani, Jerome Friedman, Trevor Hastie, Robert Tibshirani, and Jerome Friedman.
\newblock Overview of supervised learning.
\newblock {\em The elements of statistical learning: Data mining, inference, and prediction}, pages 9--41, 2009.

\bibitem{he2021debertav3}
Pengcheng He, Jianfeng Gao, and Weizhu Chen.
\newblock Debertav3: Improving deberta using electra-style pre-training with gradient-disentangled embedding sharing.
\newblock {\em arXiv preprint arXiv:2111.09543}, 2021.

\bibitem{he2020deberta}
Pengcheng He, Xiaodong Liu, Jianfeng Gao, and Weizhu Chen.
\newblock Deberta: Decoding-enhanced bert with disentangled attention.
\newblock {\em arXiv preprint arXiv:2006.03654}, 2020.

\bibitem{hoffmann2022training}
Jordan Hoffmann, Sebastian Borgeaud, Arthur Mensch, Elena Buchatskaya, Trevor Cai, Eliza Rutherford, Diego de~Las Casas, Lisa~Anne Hendricks, Johannes Welbl, Aidan Clark, et~al.
\newblock Training compute-optimal large language models.
\newblock {\em arXiv preprint arXiv:2203.15556}, 2022.

\bibitem{liger2024}
Pin-Lun Hsu, Yun Dai, Vignesh Kothapalli, Qingquan Song, Shao Tang, and Siyu Zhu.
\newblock Liger-kernel: Efficient triton kernels for llm training, 2024.

\bibitem{hu2021lora}
Edward~J Hu, Yelong Shen, Phillip Wallis, Zeyuan Allen-Zhu, Yuanzhi Li, Shean Wang, Lu~Wang, and Weizhu Chen.
\newblock Lora: Low-rank adaptation of large language models.
\newblock {\em arXiv preprint arXiv:2106.09685}, 2021.

\bibitem{huang-etal-2019-cosmos}
Lifu Huang, Ronan Le~Bras, Chandra Bhagavatula, and Yejin Choi.
\newblock Cosmos {QA}: Machine reading comprehension with contextual commonsense reasoning.
\newblock In {\em Proceedings of the 2019 Conference on Empirical Methods in Natural Language Processing and the 9th International Joint Conference on Natural Language Processing (EMNLP-IJCNLP)}, pages 2391--2401, Hong Kong, China, November 2019. Association for Computational Linguistics.

\bibitem{tokenizers}
HuggingFace.
\newblock Tokenizers: Fast state-of-the-art tokenizers optimized for research and production.
\newblock \url{https://github.com/huggingface/tokenizers}, 2019.

\bibitem{hutson2018has}
Matthew Hutson.
\newblock Has artificial intelligence become alchemy?, 2018.

\bibitem{junior2024juru}
Roseval~Malaquias Junior, Ramon Pires, Roseli Romero, and Rodrigo Nogueira.
\newblock Juru: Legal brazilian large language model from reputable sources.
\newblock {\em arXiv preprint arXiv:2403.18140}, 2024.

\bibitem{kapoor2022leakage}
Sayash Kapoor and Arvind Narayanan.
\newblock Leakage and the reproducibility crisis in ml-based science.
\newblock {\em arXiv preprint arXiv:2207.07048}, 2022.

\bibitem{kudo2018sentencepiece}
Taku Kudo and John Richardson.
\newblock Sentencepiece: A simple and language independent subword tokenizer and detokenizer for neural text processing.
\newblock {\em arXiv preprint arXiv:1808.06226}, 2018.

\bibitem{lacoste2019quantifying}
Alexandre Lacoste, Alexandra Luccioni, Victor Schmidt, and Thomas Dandres.
\newblock Quantifying the carbon emissions of machine learning.
\newblock {\em arXiv preprint arXiv:1910.09700}, 2019.

\bibitem{lai2023okapi}
Viet~Dac Lai, Chien Van~Nguyen, Nghia~Trung Ngo, Thuat Nguyen, Franck Dernoncourt, Ryan~A Rossi, and Thien~Huu Nguyen.
\newblock Okapi: Instruction-tuned large language models in multiple languages with reinforcement learning from human feedback.
\newblock {\em arXiv preprint arXiv:2307.16039}, 2023.

\bibitem{larcher2023cabrita}
Celio Larcher, Marcos Piau, Paulo Finardi, Pedro Gengo, Piero Esposito, and Vinicius Carid{\'a}.
\newblock Cabrita: closing the gap for foreign languages.
\newblock {\em arXiv preprint arXiv:2308.11878}, 2023.

\bibitem{laurenccon2022bigscience}
Hugo Lauren{\c{c}}on, Lucile Saulnier, Thomas Wang, Christopher Akiki, Albert Villanova~del Moral, Teven Le~Scao, Leandro Von~Werra, Chenghao Mou, Eduardo Gonz{\'a}lez~Ponferrada, Huu Nguyen, et~al.
\newblock The bigscience roots corpus: A 1.6 tb composite multilingual dataset.
\newblock {\em Advances in Neural Information Processing Systems}, 35:31809--31826, 2022.

\bibitem{lecun2015deep}
Yann LeCun, Yoshua Bengio, and Geoffrey Hinton.
\newblock Deep learning.
\newblock {\em nature}, 521(7553):436--444, 2015.

\bibitem{lhoest-etal-2021-datasets}
Quentin Lhoest, Albert Villanova~del Moral, Yacine Jernite, Abhishek Thakur, Patrick von Platen, Suraj Patil, Julien Chaumond, Mariama Drame, Julien Plu, Lewis Tunstall, Joe Davison, Mario {\v{S}}a{\v{s}}ko, Gunjan Chhablani, Bhavitvya Malik, Simon Brandeis, Teven Le~Scao, Victor Sanh, Canwen Xu, Nicolas Patry, Angelina McMillan-Major, Philipp Schmid, Sylvain Gugger, Cl{\'e}ment Delangue, Th{\'e}o Matussi{\`e}re, Lysandre Debut, Stas Bekman, Pierric Cistac, Thibault Goehringer, Victor Mustar, Fran{\c{c}}ois Lagunas, Alexander Rush, and Thomas Wolf.
\newblock Datasets: A community library for natural language processing.
\newblock In {\em Proceedings of the 2021 Conference on Empirical Methods in Natural Language Processing: System Demonstrations}, pages 175--184, Online and Punta Cana, Dominican Republic, November 2021. Association for Computational Linguistics.

\bibitem{li2023bactrianx}
Haonan Li, Fajri Koto, Minghao Wu, Alham~Fikri Aji, and Timothy Baldwin.
\newblock Bactrian-x : A multilingual replicable instruction-following model with low-rank adaptation, 2023.

\bibitem{li2024datacomplmsearchgenerationtraining}
Jeffrey Li, Alex Fang, Georgios Smyrnis, Maor Ivgi, Matt Jordan, Samir Gadre, Hritik Bansal, Etash Guha, Sedrick Keh, Kushal Arora, Saurabh Garg, Rui Xin, Niklas Muennighoff, Reinhard Heckel, Jean Mercat, Mayee Chen, Suchin Gururangan, Mitchell Wortsman, Alon Albalak, Yonatan Bitton, Marianna Nezhurina, Amro Abbas, Cheng-Yu Hsieh, Dhruba Ghosh, Josh Gardner, Maciej Kilian, Hanlin Zhang, Rulin Shao, Sarah Pratt, Sunny Sanyal, Gabriel Ilharco, Giannis Daras, Kalyani Marathe, Aaron Gokaslan, Jieyu Zhang, Khyathi Chandu, Thao Nguyen, Igor Vasiljevic, Sham Kakade, Shuran Song, Sujay Sanghavi, Fartash Faghri, Sewoong Oh, Luke Zettlemoyer, Kyle Lo, Alaaeldin El-Nouby, Hadi Pouransari, Alexander Toshev, Stephanie Wang, Dirk Groeneveld, Luca Soldaini, Pang~Wei Koh, Jenia Jitsev, Thomas Kollar, Alexandros~G. Dimakis, Yair Carmon, Achal Dave, Ludwig Schmidt, and Vaishaal Shankar.
\newblock Datacomp-lm: In search of the next generation of training sets for language models, 2024.

\bibitem{li2020pytorch}
Shen Li, Yanli Zhao, Rohan Varma, Omkar Salpekar, Pieter Noordhuis, Teng Li, Adam Paszke, Jeff Smith, Brian Vaughan, Pritam Damania, et~al.
\newblock Pytorch distributed: Experiences on accelerating data parallel training.
\newblock {\em arXiv preprint arXiv:2006.15704}, 2020.

\bibitem{li2023textbooks}
Yuanzhi Li, S{\'e}bastien Bubeck, Ronen Eldan, Allie Del~Giorno, Suriya Gunasekar, and Yin~Tat Lee.
\newblock Textbooks are all you need ii: phi-1.5 technical report.
\newblock {\em arXiv preprint arXiv:2309.05463}, 2023.

\bibitem{lin2021truthfulqa}
Stephanie Lin, Jacob Hilton, and Owain Evans.
\newblock Truthfulqa: Measuring how models mimic human falsehoods.
\newblock {\em arXiv preprint arXiv:2109.07958}, 2021.

\bibitem{lin2021few}
Xi~Victoria Lin, Todor Mihaylov, Mikel Artetxe, Tianlu Wang, Shuohui Chen, Daniel Simig, Myle Ott, Naman Goyal, Shruti Bhosale, Jingfei Du, et~al.
\newblock Few-shot learning with multilingual language models.
\newblock {\em arXiv preprint arXiv:2112.10668}, 2021.

\bibitem{liu2023visualinstructiontuning}
Haotian Liu, Chunyuan Li, Qingyang Wu, and Yong~Jae Lee.
\newblock Visual instruction tuning, 2023.

\bibitem{liu2019roberta}
Yinhan Liu, Myle Ott, Naman Goyal, Jingfei Du, Mandar Joshi, Danqi Chen, Omer Levy, Mike Lewis, Luke Zettlemoyer, and Veselin Stoyanov.
\newblock Roberta: A robustly optimized bert pretraining approach.
\newblock {\em arXiv preprint arXiv:1907.11692}, 2019.

\bibitem{lopes2024gl}
Ricardo Lopes, Jo{\~a}o Magalh{\~a}es, and David Semedo.
\newblock Gl$\backslash$'oria-a generative and open large language model for portuguese.
\newblock {\em arXiv preprint arXiv:2402.12969}, 2024.

\bibitem{loshchilov2017decoupled}
I~Loshchilov.
\newblock Decoupled weight decay regularization.
\newblock {\em arXiv preprint arXiv:1711.05101}, 2017.

\bibitem{lottick2019energy}
Kadan Lottick, Silvia Susai, Sorelle~A Friedler, and Jonathan~P Wilson.
\newblock Energy usage reports: Environmental awareness as part of algorithmic accountability.
\newblock {\em arXiv preprint arXiv:1911.08354}, 2019.

\bibitem{luccioni2022estimating}
AS~Luccioni, S~Viguier, and AL~Ligozat.
\newblock Estimating the carbon footprint of bloom, a 176b parameter language model, doi: 10.48550.
\newblock {\em arXiv preprint ARXIV.2211.02001}, 2022.

\bibitem{martin-etal-2020-camembert}
Louis Martin, Benjamin Muller, Pedro~Javier Ortiz~Su{\'a}rez, Yoann Dupont, Laurent Romary, {\'E}ric de~la Clergerie, Djam{\'e} Seddah, and Beno{\^\i}t Sagot.
\newblock {C}amem{BERT}: a tasty {F}rench language model.
\newblock In {\em Proceedings of the 58th Annual Meeting of the Association for Computational Linguistics}, pages 7203--7219, Online, July 2020. Association for Computational Linguistics.

\bibitem{mikolov2013efficient}
Tomas Mikolov, Kai Chen, Greg Corrado, and Jeffrey Dean.
\newblock Efficient estimation of word representations in vector space.
\newblock {\em arXiv preprint arXiv:1301.3781}, 2013.

\bibitem{mikolov2013linguistic}
Tom{\'a}{\v{s}} Mikolov, Wen-tau Yih, and Geoffrey Zweig.
\newblock Linguistic regularities in continuous space word representations.
\newblock In {\em Proceedings of the 2013 conference of the north american chapter of the association for computational linguistics: Human language technologies}, pages 746--751, 2013.

\bibitem{misra2020self}
Ishan Misra and Laurens van~der Maaten.
\newblock Self-supervised learning of pretext-invariant representations.
\newblock In {\em Proceedings of the IEEE/CVF conference on computer vision and pattern recognition}, pages 6707--6717, 2020.

\bibitem{mitra2024orca}
Arindam Mitra, Hamed Khanpour, Corby Rosset, and Ahmed Awadallah.
\newblock Orca-math: Unlocking the potential of slms in grade school math.
\newblock {\em arXiv preprint arXiv:2402.14830}, 2024.

\bibitem{chenghao_mou_2023_8364980}
Chenghao Mou, Chris Ha, Kenneth Enevoldsen, and Peiyuan Liu.
\newblock Chenghaomou/text-dedup: Reference snapshot, September 2023.

\bibitem{muennighoff2023scaling}
Niklas Muennighoff, Alexander~M Rush, Boaz Barak, Teven~Le Scao, Aleksandra Piktus, Nouamane Tazi, Sampo Pyysalo, Thomas Wolf, and Colin Raffel.
\newblock Scaling data-constrained language models.
\newblock {\em arXiv preprint arXiv:2305.16264}, 2023.

\bibitem{nadkarni2011natural}
Prakash~M Nadkarni, Lucila Ohno-Machado, and Wendy~W Chapman.
\newblock Natural language processing: an introduction.
\newblock {\em Journal of the American Medical Informatics Association}, 18(5):544--551, 2011.

\bibitem{nguyen2022quality}
Thao Nguyen, Gabriel Ilharco, Mitchell Wortsman, Sewoong Oh, and Ludwig Schmidt.
\newblock Quality not quantity: On the interaction between dataset design and robustness of clip.
\newblock {\em Advances in Neural Information Processing Systems}, 35:21455--21469, 2022.

\bibitem{nguyen2023culturax}
Thuat Nguyen, Chien~Van Nguyen, Viet~Dac Lai, Hieu Man, Nghia~Trung Ngo, Franck Dernoncourt, Ryan~A. Rossi, and Thien~Huu Nguyen.
\newblock Culturax: A cleaned, enormous, and multilingual dataset for large language models in 167 languages, 2023.

\bibitem{niklaus2023multilegalpile}
Joel Niklaus, Veton Matoshi, Matthias St{\"u}rmer, Ilias Chalkidis, and Daniel~E Ho.
\newblock Multilegalpile: A 689gb multilingual legal corpus.
\newblock {\em arXiv preprint arXiv:2306.02069}, 2023.

\bibitem{ortiz-suarez-etal-2020-monolingual}
Pedro~Javier Ortiz~Su{'a}rez, Laurent Romary, and Benoit Sagot.
\newblock A monolingual approach to contextualized word embeddings for mid-resource languages.
\newblock In {\em Proceedings of the 58th Annual Meeting of the Association for Computational Linguistics}, pages 1703--1714, Online, July 2020. Association for Computational Linguistics.

\bibitem{OrtizSuarezSagotRomary2019}
Pedro~Javier {Ortiz Su{'a}rez}, Benoit Sagot, and Laurent Romary.
\newblock Asynchronous pipelines for processing huge corpora on medium to low resource infrastructures.
\newblock Proceedings of the Workshop on Challenges in the Management of Large Corpora (CMLC-7) 2019. Cardiff, 22nd July 2019, pages 9 -- 16, Mannheim, 2019. Leibniz-Institut f{"u}r Deutsche Sprache.

\bibitem{otter2020survey}
Daniel~W Otter, Julian~R Medina, and Jugal~K Kalita.
\newblock A survey of the usages of deep learning for natural language processing.
\newblock {\em IEEE transactions on neural networks and learning systems}, 32(2):604--624, 2020.

\bibitem{ouyang2022training}
Long Ouyang, Jeffrey Wu, Xu~Jiang, Diogo Almeida, Carroll Wainwright, Pamela Mishkin, Chong Zhang, Sandhini Agarwal, Katarina Slama, Alex Ray, et~al.
\newblock Training language models to follow instructions with human feedback.
\newblock {\em Advances in neural information processing systems}, 35:27730--27744, 2022.

\bibitem{overwijk2022clueweb22}
Arnold Overwijk, Chenyan Xiong, and Jamie Callan.
\newblock Clueweb22: 10 billion web documents with rich information.
\newblock In {\em Proceedings of the 45th International ACM SIGIR Conference on Research and Development in Information Retrieval}, pages 3360--3362, 2022.

\bibitem{paperno2016lambada}
Denis Paperno, Germ{\'a}n Kruszewski, Angeliki Lazaridou, Quan~Ngoc Pham, Raffaella Bernardi, Sandro Pezzelle, Marco Baroni, Gemma Boleda, and Raquel Fern{\'a}ndez.
\newblock The lambada dataset: Word prediction requiring a broad discourse context.
\newblock {\em arXiv preprint arXiv:1606.06031}, 2016.

\bibitem{parmar2024nemotron}
Jupinder Parmar, Shrimai Prabhumoye, Joseph Jennings, Mostofa Patwary, Sandeep Subramanian, Dan Su, Chen Zhu, Deepak Narayanan, Aastha Jhunjhunwala, Ayush Dattagupta, et~al.
\newblock Nemotron-4 15b technical report.
\newblock {\em arXiv preprint arXiv:2402.16819}, 2024.

\bibitem{paszke2019pytorch}
Adam Paszke, Sam Gross, Francisco Massa, Adam Lerer, James Bradbury, Gregory Chanan, Trevor Killeen, Zeming Lin, Natalia Gimelshein, Luca Antiga, et~al.
\newblock Pytorch: An imperative style, high-performance deep learning library.
\newblock {\em Advances in neural information processing systems}, 32, 2019.

\bibitem{patterson2022carbon}
David Patterson, Joseph Gonzalez, Urs H{\"o}lzle, Quoc Le, Chen Liang, Lluis-Miquel Munguia, Daniel Rothchild, David~R So, Maud Texier, and Jeff Dean.
\newblock The carbon footprint of machine learning training will plateau, then shrink.
\newblock {\em Computer}, 55(7):18--28, 2022.

\bibitem{penedo2024finewebdatasetsdecantingweb}
Guilherme Penedo, Hynek Kydlíček, Loubna~Ben allal, Anton Lozhkov, Margaret Mitchell, Colin Raffel, Leandro~Von Werra, and Thomas Wolf.
\newblock The fineweb datasets: Decanting the web for the finest text data at scale, 2024.

\bibitem{piau2024ptt5}
Marcos Piau, Roberto Lotufo, and Rodrigo Nogueira.
\newblock ptt5-v2: A closer look at continued pretraining of t5 models for the portuguese language.
\newblock {\em arXiv preprint arXiv:2406.10806}, 2024.

\bibitem{pires2023sabi}
Ramon Pires, Hugo Abonizio, Thales Rog{\'e}rio, and Rodrigo Nogueira.
\newblock Sabi$\backslash$'a: Portuguese large language models.
\newblock {\em arXiv preprint arXiv:2304.07880}, 2023.

\bibitem{polo2021legalnlp}
Felipe~Maia Polo, Gabriel Caiaffa~Floriano Mendon{\c{c}}a, Kau{\^e} Capellato~J Parreira, Lucka Gianvechio, Peterson Cordeiro, Jonathan~Batista Ferreira, Leticia Maria~Paz de~Lima, Ant{\^o}nio Carlos do~Amaral Maia, and Renato Vicente.
\newblock Legalnlp--natural language processing methods for the brazilian legal language.
\newblock {\em arXiv preprint arXiv:2110.15709}, 2021.

\bibitem{prince2023understanding}
Simon~JD Prince.
\newblock {\em Understanding deep learning}.
\newblock MIT press, 2023.

\bibitem{radford2019language}
Alec Radford, Jeffrey Wu, Rewon Child, David Luan, Dario Amodei, Ilya Sutskever, et~al.
\newblock Language models are unsupervised multitask learners.
\newblock {\em OpenAI blog}, 1(8):9, 2019.

\bibitem{rae2021scaling}
Jack~W Rae, Sebastian Borgeaud, Trevor Cai, Katie Millican, Jordan Hoffmann, Francis Song, John Aslanides, Sarah Henderson, Roman Ring, Susannah Young, et~al.
\newblock Scaling language models: Methods, analysis \& insights from training gopher.
\newblock {\em arXiv preprint arXiv:2112.11446}, 2021.

\bibitem{rafailov2024direct}
Rafael Rafailov, Archit Sharma, Eric Mitchell, Christopher~D Manning, Stefano Ermon, and Chelsea Finn.
\newblock Direct preference optimization: Your language model is secretly a reward model.
\newblock {\em Advances in Neural Information Processing Systems}, 36, 2024.

\bibitem{raffel2020exploring}
Colin Raffel, Noam Shazeer, Adam Roberts, Katherine Lee, Sharan Narang, Michael Matena, Yanqi Zhou, Wei Li, and Peter~J Liu.
\newblock Exploring the limits of transfer learning with a unified text-to-text transformer.
\newblock {\em Journal of machine learning research}, 21(140):1--67, 2020.

\bibitem{real2020assin}
Livy Real, Erick Fonseca, and Hugo~Goncalo Oliveira.
\newblock The assin 2 shared task: a quick overview.
\newblock In {\em International Conference on Computational Processing of the Portuguese Language}, pages 406--412. Springer, 2020.

\bibitem{reimers2019sentence}
Nils Reimers and Iryna Gurevych.
\newblock Sentence-bert: Sentence embeddings using siamese bert-networks.
\newblock {\em arXiv preprint arXiv:1908.10084}, 2019.

\bibitem{rodrigues2023advancing}
Jo{\~a}o Rodrigues, Lu{\'\i}s Gomes, Jo{\~a}o Silva, Ant{\'o}nio Branco, Rodrigo Santos, Henrique~Lopes Cardoso, and Tom{\'a}s Os{\'o}rio.
\newblock Advancing neural encoding of portuguese with transformer albertina pt.
\newblock {\em arXiv preprint arXiv:2305.06721}, 2023.

\bibitem{rodrigues2022petrobert}
Rafael~BM Rodrigues, Pedro~IM Privatto, Gustavo~Jos{\'e} de~Sousa, Rafael~P Murari, Luis~CS Afonso, Jo{\~a}o~P Papa, Daniel~CG Pedronette, Ivan~R Guilherme, Stephan~R Perrout, and Aliel~F Riente.
\newblock Petrobert: a domain adaptation language model for oil and gas applications in portuguese.
\newblock In {\em International Conference on Computational Processing of the Portuguese Language}, pages 101--109. Springer, 2022.

\bibitem{faquad-nli-2029}
Ruan~Chaves Rodrigues.
\newblock Faquad-nli.
\newblock \url{https://huggingface.co/datasets/ruanchaves/faquad-nli}, 2024.

\bibitem{rubel2020biobertpt}
Elisa~Terumi Rubel~Schneider, Jo{\~a}o~Vitor Andrioli~de Souza, Julien Knafou, Lucas~ES Oliveira, Yohan~B Gumiel, Lucas~FA de~Oliveira, Douglas Teodoro, Emerson~Cabrera Paraiso, Claudia Moro, et~al.
\newblock Biobertpt: a portuguese neural language model for clinical named entity recognition.
\newblock In {\em Proceedings of the 3rd Clinical Natural Language Processing Workshop}. 19 November 2020, 2020.

\bibitem{santos2018blogset}
Henrique Santos, Vinicius Woloszyn, and Renata Vieira.
\newblock Blogset-br: A brazilian portuguese blog corpus.
\newblock In {\em Proceedings of the Eleventh International Conference on Language Resources and Evaluation (LREC 2018)}, 2018.

\bibitem{santos2024advancing}
Rodrigo Santos, Jo{\~a}o Silva, Lu{\'\i}s Gomes, Jo{\~a}o Rodrigues, and Ant{\'o}nio Branco.
\newblock Advancing generative ai for portuguese with open decoder gerv$\backslash$'asio pt.
\newblock {\em arXiv preprint arXiv:2402.18766}, 2024.

\bibitem{schmidt2024tokenization}
Craig~W Schmidt, Varshini Reddy, Haoran Zhang, Alec Alameddine, Omri Uzan, Yuval Pinter, and Chris Tanner.
\newblock Tokenization is more than compression.
\newblock {\em arXiv preprint arXiv:2402.18376}, 2024.

\bibitem{schneider2021gpt}
Elisa Terumi~Rubel Schneider, Joao Vitor~Andrioli de~Souza, Yohan~Bonescki Gumiel, Claudia Moro, and Emerson~Cabrera Paraiso.
\newblock A gpt-2 language model for biomedical texts in portuguese.
\newblock In {\em 2021 IEEE 34th international symposium on computer-based medical systems (CBMS)}, pages 474--479. IEEE, 2021.

\bibitem{shallue2019measuring}
Christopher~J Shallue, Jaehoon Lee, Joseph Antognini, Jascha Sohl-Dickstein, Roy Frostig, and George~E Dahl.
\newblock Measuring the effects of data parallelism on neural network training.
\newblock {\em Journal of Machine Learning Research}, 20(112):1--49, 2019.

\bibitem{shazeer2020glu}
Noam Shazeer.
\newblock Glu variants improve transformer.
\newblock {\em arXiv preprint arXiv:2002.05202}, 2020.

\bibitem{shliazhko2022mgpt}
Oleh Shliazhko, Alena Fenogenova, Maria Tikhonova, Vladislav Mikhailov, Anastasia Kozlova, and Tatiana Shavrina.
\newblock mgpt: Few-shot learners go multilingual.
\newblock {\em arXiv preprint arXiv:2204.07580}, 2022.

\bibitem{ENEM-Challenge}
Igor~Cataneo Silveira and Denis~Deratani Mau\'a.
\newblock University entrance exam as a guiding test for artificial intelligence.
\newblock In {\em Proceedings of the 6th Brazilian Conference on Intelligent Systems}, BRACIS, pages 426--431, 2017.

\bibitem{silveira2023legalbert}
Raquel Silveira, Caio Ponte, Vitor Almeida, Vl{\'a}dia Pinheiro, and Vasco Furtado.
\newblock Legalbert-pt: A pretrained language model for the brazilian portuguese legal domain.
\newblock In {\em Brazilian Conference on Intelligent Systems}, pages 268--282. Springer, 2023.

\bibitem{singh2024ayadatasetopenaccesscollection}
Shivalika Singh, Freddie Vargus, Daniel Dsouza, Börje~F. Karlsson, Abinaya Mahendiran, Wei-Yin Ko, Herumb Shandilya, Jay Patel, Deividas Mataciunas, Laura OMahony, Mike Zhang, Ramith Hettiarachchi, Joseph Wilson, Marina Machado, Luisa~Souza Moura, Dominik Krzemiński, Hakimeh Fadaei, Irem Ergün, Ifeoma Okoh, Aisha Alaagib, Oshan Mudannayake, Zaid Alyafeai, Vu~Minh Chien, Sebastian Ruder, Surya Guthikonda, Emad~A. Alghamdi, Sebastian Gehrmann, Niklas Muennighoff, Max Bartolo, Julia Kreutzer, Ahmet Üstün, Marzieh Fadaee, and Sara Hooker.
\newblock Aya dataset: An open-access collection for multilingual instruction tuning, 2024.

\bibitem{dolma}
Luca Soldaini, Rodney Kinney, Akshita Bhagia, Dustin Schwenk, David Atkinson, Russell Authur, Ben Bogin, Khyathi Chandu, Jennifer Dumas, Yanai Elazar, Valentin Hofmann, Ananya~Harsh Jha, Sachin Kumar, Li~Lucy, Xinxi Lyu, Nathan Lambert, Ian Magnusson, Jacob Morrison, Niklas Muennighoff, Aakanksha Naik, Crystal Nam, Matthew~E. Peters, Abhilasha Ravichander, Kyle Richardson, Zejiang Shen, Emma Strubell, Nishant Subramani, Oyvind Tafjord, Pete Walsh, Luke Zettlemoyer, Noah~A. Smith, Hannaneh Hajishirzi, Iz~Beltagy, Dirk Groeneveld, Jesse Dodge, and Kyle Lo.
\newblock {Dolma: an Open Corpus of Three Trillion Tokens for Language Model Pretraining Research}.
\newblock {\em arXiv preprint}, 2024.

\bibitem{sousa2019iudicium}
A~Willian Sousa and Marcos~Didonet Del~Fabro.
\newblock Iudicium textum dataset uma base de textos jur{\i}dicos para nlp.
\newblock In {\em XXXIV Simp{\'o}sio Brasileiro de Banco de Dados: Dataset Showcase Workshop, SBBD}, pages 1--11, 2019.

\bibitem{souza2020bertimbau}
F{\'a}bio Souza, Rodrigo Nogueira, and Roberto Lotufo.
\newblock Bertimbau: pretrained bert models for brazilian portuguese.
\newblock In {\em Intelligent Systems: 9th Brazilian Conference, BRACIS 2020, Rio Grande, Brazil, October 20--23, 2020, Proceedings, Part I 9}, pages 403--417. Springer, 2020.

\bibitem{srivastava2024lolaopensourcemassively}
Nikit Srivastava, Denis Kuchelev, Tatiana~Moteu Ngoli, Kshitij Shetty, Michael Roeder, Diego Moussallem, Hamada Zahera, and Axel-Cyrille~Ngonga Ngomo.
\newblock Lola -- an open-source massively multilingual large language model, 2024.

\bibitem{strubell2019energy}
Emma Strubell, Ananya Ganesh, and Andrew McCallum.
\newblock Energy and policy considerations for deep learning in nlp.
\newblock {\em arXiv preprint arXiv:1906.02243}, 2019.

\bibitem{su2021roformer}
Jianlin Su, Yu~Lu, Shengfeng Pan, Bo~Wen, and Yunfeng Liu.
\newblock Roformer: Enhanced transformer with rotary position embedding. corr abs/2104.09864 (2021).
\newblock {\em arXiv preprint arXiv:2104.09864}, 2021.

\bibitem{sun2021ernie}
Yu~Sun, Shuohuan Wang, Shikun Feng, Siyu Ding, Chao Pang, Junyuan Shang, Jiaxiang Liu, Xuyi Chen, Yanbin Zhao, Yuxiang Lu, et~al.
\newblock Ernie 3.0: Large-scale knowledge enhanced pre-training for language understanding and generation.
\newblock {\em arXiv preprint arXiv:2107.02137}, 2021.

\bibitem{sutskever2014sequence}
Ilya Sutskever, Oriol Vinyals, and Quoc~V Le.
\newblock Sequence to sequence learning with neural networks.
\newblock {\em Advances in neural information processing systems}, 27, 2014.

\bibitem{tan20241}
Calvin Tan and Jerome Wang.
\newblock 1.5-pints technical report: Pretraining in days, not months--your language model thrives on quality data.
\newblock {\em arXiv preprint arXiv:2408.03506}, 2024.

\bibitem{taori2023alpaca}
Rohan Taori, Ishaan Gulrajani, Tianyi Zhang, Yann Dubois, Xuechen Li, Carlos Guestrin, Percy Liang, and Tatsunori~B Hashimoto.
\newblock Alpaca: A strong, replicable instruction-following model.
\newblock {\em Stanford Center for Research on Foundation Models. https://crfm. stanford. edu/2023/03/13/alpaca. html}, 3(6):7, 2023.

\bibitem{touvron2023llama1}
Hugo Touvron, Thibaut Lavril, Gautier Izacard, Xavier Martinet, Marie-Anne Lachaux, Timoth{\'e}e Lacroix, Baptiste Rozi{\`e}re, Naman Goyal, Eric Hambro, Faisal Azhar, et~al.
\newblock Llama: Open and efficient foundation language models.
\newblock {\em arXiv preprint arXiv:2302.13971}, 2023.

\bibitem{touvron2023llama2}
Hugo Touvron, Louis Martin, Kevin Stone, Peter Albert, Amjad Almahairi, Yasmine Babaei, Nikolay Bashlykov, Soumya Batra, Prajjwal Bhargava, Shruti Bhosale, et~al.
\newblock Llama 2: Open foundation and fine-tuned chat models.
\newblock {\em arXiv preprint arXiv:2307.09288}, 2023.

\bibitem{van2021sustainable}
Aimee Van~Wynsberghe.
\newblock Sustainable ai: Ai for sustainability and the sustainability of ai.
\newblock {\em AI and Ethics}, 1(3):213--218, 2021.

\bibitem{vargas2022hatebr}
Francielle Vargas, Isabelle Carvalho, Fabiana~Rodrigues de~G{\'o}es, Thiago Pardo, and Fabr{\'\i}cio Benevenuto.
\newblock Hatebr: A large expert annotated corpus of brazilian instagram comments for offensive language and hate speech detection.
\newblock In {\em Proceedings of the Thirteenth Language Resources and Evaluation Conference}, pages 7174--7183, 2022.

\bibitem{vaswani2017attention}
Ashish Vaswani, Noam Shazeer, Niki Parmar, Jakob Uszkoreit, Llion Jones, Aidan~N Gomez, {\L}ukasz Kaiser, and Illia Polosukhin.
\newblock Attention is all you need.
\newblock {\em Advances in neural information processing systems}, 30, 2017.

\bibitem{viegas2023jurisbert}
Charles~FO Viegas, Bruno~C Costa, and Renato~P Ishii.
\newblock Jurisbert: a new approach that converts a classification corpus into an sts one.
\newblock In {\em International Conference on Computational Science and Its Applications}, pages 349--365. Springer, 2023.

\bibitem{virtanen2019multilingual}
Antti Virtanen, Jenna Kanerva, Rami Ilo, Jouni Luoma, Juhani Luotolahti, Tapio Salakoski, Filip Ginter, and Sampo Pyysalo.
\newblock Multilingual is not enough: Bert for finnish.
\newblock {\em arXiv preprint arXiv:1912.07076}, 2019.

\bibitem{vonwerra2022trl}
Leandro von Werra, Younes Belkada, Lewis Tunstall, Edward Beeching, Tristan Thrush, Nathan Lambert, Shengyi Huang, Kashif Rasul, and Quentin Gallouédec.
\newblock Trl: Transformer reinforcement learning.
\newblock \url{https://github.com/huggingface/trl}, 2020.

\bibitem{wagner2018brwac}
Jorge~A Wagner~Filho, Rodrigo Wilkens, Marco Idiart, and Aline Villavicencio.
\newblock The brwac corpus: A new open resource for brazilian portuguese.
\newblock In {\em Proceedings of the Eleventh International Conference on Language Resources and Evaluation (LREC 2018)}, 2018.

\bibitem{gptj}
Ben Wang and Aran Komatsuzaki.
\newblock {GPT-J-6B: A 6 Billion Parameter Autoregressive Language Model}.
\newblock \url{https://github.com/kingoflolz/mesh-transformer-jax}, 2021.

\bibitem{wang2024finetuned}
Weizhi Wang, Khalil Mrini, Linjie Yang, Sateesh Kumar, Yu~Tian, Xifeng Yan, and Heng Wang.
\newblock Finetuned multimodal language models are high-quality image-text data filters.
\newblock {\em arXiv preprint arXiv:2403.02677}, 2024.

\bibitem{wei2023polylm}
Xiangpeng Wei, Haoran Wei, Huan Lin, Tianhao Li, Pei Zhang, Xingzhang Ren, Mei Li, Yu~Wan, Zhiwei Cao, Binbin Xie, et~al.
\newblock Polylm: An open source polyglot large language model.
\newblock {\em arXiv preprint arXiv:2307.06018}, 2023.

\bibitem{wenzek-etal-2020-ccnet}
Guillaume Wenzek, Marie-Anne Lachaux, Alexis Conneau, Vishrav Chaudhary, Francisco Guzm{\'a}n, Armand Joulin, and Edouard Grave.
\newblock {CCN}et: Extracting high quality monolingual datasets from web crawl data.
\newblock In {\em Proceedings of the 12th Language Resources and Evaluation Conference}, pages 4003--4012, Marseille, France, May 2020. European Language Resources Association.

\bibitem{wikidump}
{Wikimedia Foundation}.
\newblock {Wikimedia Downloads}.
\newblock \url{https://dumps.wikimedia.org}, 2024.

\bibitem{workshop2022bloom}
BigScience Workshop, Teven~Le Scao, Angela Fan, Christopher Akiki, Ellie Pavlick, Suzana Ili{\'c}, Daniel Hesslow, Roman Castagn{\'e}, Alexandra~Sasha Luccioni, Fran{\c{c}}ois Yvon, et~al.
\newblock Bloom: A 176b-parameter open-access multilingual language model.
\newblock {\em arXiv preprint arXiv:2211.05100}, 2022.

\bibitem{xue2023repeat}
Fuzhao Xue, Yao Fu, Wangchunshu Zhou, Zangwei Zheng, and Yang You.
\newblock To repeat or not to repeat: Insights from scaling llm under token-crisis.
\newblock {\em arXiv preprint arXiv:2305.13230}, 2023.

\bibitem{xue2024openmoe}
Fuzhao Xue, Zian Zheng, Yao Fu, Jinjie Ni, Zangwei Zheng, Wangchunshu Zhou, and Yang You.
\newblock Openmoe: An early effort on open mixture-of-experts language models.
\newblock {\em arXiv preprint arXiv:2402.01739}, 2024.

\bibitem{xue2020mt5}
Linting Xue, Noah Constant, Adam Roberts, Mihir Kale, Rami Al-Rfou, Aditya Siddhant, Aditya Barua, and Colin Raffel.
\newblock mt5: A massively multilingual pre-trained text-to-text transformer.
\newblock {\em arXiv preprint arXiv:2010.11934}, 2020.

\bibitem{young2024yi}
Alex Young, Bei Chen, Chao Li, Chengen Huang, Ge~Zhang, Guanwei Zhang, Heng Li, Jiangcheng Zhu, Jianqun Chen, Jing Chang, et~al.
\newblock Yi: Open foundation models by 01. ai.
\newblock {\em arXiv preprint arXiv:2403.04652}, 2024.

\bibitem{zellers2019hellaswag}
Rowan Zellers, Ari Holtzman, Yonatan Bisk, Ali Farhadi, and Yejin Choi.
\newblock Hellaswag: Can a machine really finish your sentence?
\newblock {\em arXiv preprint arXiv:1905.07830}, 2019.

\bibitem{zhang2019root}
Biao Zhang and Rico Sennrich.
\newblock Root mean square layer normalization.
\newblock {\em Advances in Neural Information Processing Systems}, 32, 2019.

\bibitem{zhang2024tinyllama}
Peiyuan Zhang, Guangtao Zeng, Tianduo Wang, and Wei Lu.
\newblock Tinyllama: An open-source small language model.
\newblock {\em arXiv preprint arXiv:2401.02385}, 2024.

\bibitem{zhang2022opt}
Susan Zhang, Stephen Roller, Naman Goyal, Mikel Artetxe, Moya Chen, Shuohui Chen, Christopher Dewan, Mona Diab, Xian Li, Xi~Victoria Lin, et~al.
\newblock Opt: Open pre-trained transformer language models.
\newblock {\em arXiv preprint arXiv:2205.01068}, 2022.

\bibitem{ustun2024ayamodelinstructionfinetuned}
Ahmet Üstün, Viraat Aryabumi, Zheng-Xin Yong, Wei-Yin Ko, Daniel D'souza, Gbemileke Onilude, Neel Bhandari, Shivalika Singh, Hui-Lee Ooi, Amr Kayid, Freddie Vargus, Phil Blunsom, Shayne Longpre, Niklas Muennighoff, Marzieh Fadaee, Julia Kreutzer, and Sara Hooker.
\newblock Aya model: An instruction finetuned open-access multilingual language model, 2024.

\end{thebibliography}

\end{document}